\documentclass{article} 
\PassOptionsToPackage{dvipsnames}{xcolor}

\usepackage{amsmath,amsfonts,bm}

\def\eqref#1{equation~\ref{#1}}

\def\1{\bm{1}}

\DeclareMathAlphabet{\mathsfit}{\encodingdefault}{\sfdefault}{m}{sl}
\SetMathAlphabet{\mathsfit}{bold}{\encodingdefault}{\sfdefault}{bx}{n}

\usepackage{url}
\usepackage[dvipsnames]{xcolor}
\usepackage{mathtools}
\usepackage{amsmath}
\usepackage{amsthm}
\usepackage{enumerate}
\usepackage{graphicx}
\usepackage[export]{adjustbox}
\usepackage{booktabs}
\usepackage{tikz}
\usepackage{tikzsymbols}
\usetikzlibrary{calc}
\usepackage{wasysym}
\usepackage{pgfplots}
\usepackage{listings}

\usepackage[labelformat=simple]{subcaption}

\usepackage[normalem]{ulem}
\usepackage{doi}
\useunder{\uline}{\ul}{}
\usepackage[toc,page]{appendix}
\usepackage{paralist}
\usepackage{cleveref}
\usepackage{scalerel}
\usepackage{multirow}
\usepackage{wrapfig}
\usepackage[frozencache]{minted}
\usepackage{makecell}
\usepackage{enumitem}
\usepackage{xspace}
\usepackage{hyperref}

\newcommand{\omer}[1]{{\color{blue}omer:[#1]}}
\newcommand{\uri}[1]{{\color{OliveGreen}Uri:[#1]}}
\newcommand{\eran}[1]{{\color{purple}eran:[#1]}}

\renewcommand{\omer}[1]{}
\renewcommand{\uri}[1]{}
\renewcommand{\eran}[1]{}

\newcommand{\greenbox}[1]{\setlength{\fboxrule}{1pt}\fcolorbox{green}{green!5}{#1}}
\definecolor{lightRed}{HTML}{F8CECC}

\newcommand{\scode}[1]{{\small \texttt{#1}}}
\renewcommand\cite{\citep}

\newcommand{\parent}{\pi\left(a_t\right)}
\newcommand{\fulltask}{any-code completion}
\newcommand{\Fulltask}{Any-code completion}
\newcommand{\aeg}{Any-Code Completion}
\newcommand{\tasktitle}{\aeg}
\newcommand{\restrictgen}{restricted completion}

\makeatletter
\newcommand{\oset}[3][0ex]{%
  \mathrel{\mathop{#3}\limits^{
    \vbox to#1{\kern-2\ex@
    \hbox{$\scriptstyle#2$}\vss}}}}
\makeatother

\newcommand{\fpath}{\oset[-.2ex]{\leadsto}{f}}

\definecolor{lightbluebackground}{HTML}{DAE8FC}
\definecolor{lightblueborder}{HTML}{6C8EBF}
\newcommand{\cfbox}[2]{%
    \colorlet{currentcolor}{.}%
    {\color{#1}%
    \fbox{\color{currentcolor}#2}}%
}

\newcommand\para[1]{\noindent \textbf{#1}}
\crefname{algocf}{Algo.}{algs.}
\Crefname{algocf}{Algorithm}{Algorithms}
\usepackage[compact]{titlesec}

\usepackage[accepted]{icml2020}
\newcommand{\ourtitle}{Structural Language Models of Code}
\icmltitlerunning{\ourtitle}

\begin{document}

\twocolumn[
\icmltitle{\ourtitle}

\icmlsetsymbol{equal}{*}
\begin{icmlauthorlist}
\icmlauthor{Uri Alon}{technion}
\icmlauthor{Roy Sadaka}{technion}
\icmlauthor{Omer Levy}{tau,fair}
\icmlauthor{Eran Yahav}{technion}
\end{icmlauthorlist}

\icmlaffiliation{technion}{Technion, Israel}
\icmlaffiliation{fair}{Facebook AI Research}
\icmlaffiliation{tau}{Tel Aviv University}

\icmlcorrespondingauthor{Uri Alon}{urialon@cs.technion.ac.il}
\icmlcorrespondingauthor{Roy Sadaka}{roysadaka@gmail.com}
\icmlcorrespondingauthor{Omer Levy}{omerlevy@gmail.com}
\icmlcorrespondingauthor{Eran Yahav}{yahave@cs.technion.ac.il}

\icmlkeywords{Machine Learning, ICML}

\vskip 0.3in
]
\printAffiliationsAndNotice{}

\begin{abstract}
We address the problem of \emph{\fulltask{}} -- 
generating a missing piece of source code in a given program
without any restriction on the vocabulary or structure.
We introduce a new approach to \fulltask{} that leverages the strict syntax of programming languages to model a code snippet as a tree -- \emph{structural language modeling} (SLM).
SLM estimates the probability of the program's abstract syntax tree (AST) by decomposing it into a product of conditional probabilities over its nodes.
We present a neural model that computes these conditional probabilities by considering all AST paths leading to a target node.
Unlike previous 
techniques that have severely restricted the kinds of expressions that can be generated in this task, our approach can generate arbitrary code in any programming language.
Our model significantly outperforms both seq2seq and a variety of structured approaches in generating Java and C\# code.
Our code, data, and trained models are available at {\small{\url{http://github.com/tech-srl/slm-code-generation/}}}.
An online demo is available at {\small{\url{http://AnyCodeGen.org}}}.
\end{abstract}

\section{Introduction}
\label{Intro}

Code completion is the problem of generating code given its surrounding code as context. In its most general form, this problem is extremely challenging as it requires reasoning over an unbounded number of syntactic structures and user-defined symbols.
Previous approaches have avoided this issue by limiting the generation problem:
program synthesis approaches 
are often tailored to domain-specific languages~\cite{gulwani2011automating, polozov2015flashmeta, devlin2017robustfill,ellis2019write},
while other recent approaches generate code in general languages like Java and C\#, but severely restrict the syntax, vocabulary, domain, or nature of the generated programs~\cite{murali2018bayou, brockschmidt2018generative, young2019learning}. 

\newcommand{\specialcell}[2][l]{%
  \begin{tabular}[#1]{@{}l@{}}#2\end{tabular}}
\begin{figure*}[t]
\centering
\hspace{-0.6mm}
\begin{minipage}{0.47\textwidth}
\begin{minted}[fontsize=\footnotesize,stripnl=false,frame=single,framesep=2pt,escapeinside=||]{java}
public static Path[] stat2Paths(
    FileStatus[] stats) {
  if (stats == null) return null;
  Path[] ret = new Path[stats.length];
  for (int i = 0; i < stats.length; ++i){
    ret[i] = |\greenbox{\phantom{stats[i].getPath()}}|;
  }
  return ret;
}
\end{minted}
\end{minipage}
\hspace{6mm}
\begin{minipage}{0.45\textwidth}
 \centering
\begin{minted}[fontsize=\footnotesize, frame=single,framesep=2pt,escapeinside=||]{csharp}
public static string Camelize(
    this string input)
{
    var word = input.Pascalize();
    return word.Length > 0 ? 
        |\greenbox{\phantom{word.Substring(0, 1)}}|.ToLower() 
            + word.Substring(1) 
        : word;
}
\end{minted}
\end{minipage}

 \begin{minipage}{0.49\textwidth}
 \begin{subfigure}[t]{1\textwidth}
 \footnotesize
 \centering
 \begin{tabular}{crl}
 \toprule
\multicolumn{2}{l}{True ref (Java):}  &\scode{stats[i].getPath()} \\
 \midrule
 \multirow{5}{*}{\makecell[l]{SLM\\top-5:}} & (25.2\%) & \scode{\textbf{stats[i].getPath()}}  \\
 & (3.3\%)  & \scode{Path(stats[i])}\\
 & (2.5\%)  & \scode{new Path(stats[i], charset)} \\
 & (1.7\%)  & \scode{stat(stats[i], ret)} \\
 & (0.8\%)  & \scode{new Path(stats[i])} \\
 \bottomrule
 \end{tabular}
 \caption{}
 \label{fig:task_example_a}
 \end{subfigure}
 \end{minipage}
 \begin{minipage}{0.49\textwidth}
  \hspace{1mm}
 \begin{subfigure}[t]{1\textwidth}
 \footnotesize
 \centering
 \begin{tabular}{crl}
 \toprule
 \multicolumn{2}{l}{True ref (C\#): } & \scode{word.Substring(0, 1)} \\
 \midrule
 \multirow{5}{*}{\makecell[l]{SLM\\top-5:}} & (14.1\%) & \scode{\textbf{word.Substring(0, 1)}	} \\
  & (8.2\%)  & \scode{word.trim()} \\
  & (5.8\%)  & \scode{word.Substring(1)} \\
  & (2.4\%)  & \scode{input.Substring(0, 1)} \\
  & (1.9\%)  & \scode{wordValue.Substring(0, 1)} \\
\bottomrule
 \end{tabular}
 \caption{}
 \label{fig:task_example_b}
 \end{subfigure}
 \end{minipage}

\caption{Examples from the Java (left) and C\# (right) test sets. The highlighted expression in each example is the target $p$, which our models correctly generated from the rest of the snippet. Additional and larger examples can be found in \Cref{appendix:java_examples,appendix:csharp_examples}.}
\label{fig:task_example}
\end{figure*}

We introduce the task of \emph{\fulltask} -- generating code in a general-purpose programming language without any restriction on its vocabulary or structure.
Specifically, we focus on generating code in context:  
given a program $\mathcal{P}$ and some part of the program $p$, the task is to predict $p$ from the rest of the program $\mathcal{P^{-}}$$=$$\mathcal{P}$$\setminus$$p$.
\Fulltask{} thus generalizes the restricted completion task of \citet{brockschmidt2018generative}, in which the target code contained only primitive types (e.g., \scode{int} and \scode{string}) and excluded user-defined functions.
\Cref{fig:task_example} shows two \fulltask{} examples.
\omer{Do we have an example where the missing piece is an entire statement? Isn't this another aspect where we're much more daring than previous work?} \uri{I do have, but in another model/sub-dataset which I did not perform all the experiment on...}

In related tasks such as semantic parsing \cite{dong2018coarse, yu2018typesql, iyer2019learning}, natural-language-to-code \cite{bimodal15,iyer2018mapping}, and edit-to-code \cite{yin2018learning, zhao2019neural}, models must use separate encoders and decoders because of the different modalities of the input (e.g. natural language text) and the output (code). In contrast, we leverage the fact that our input and output are of the \emph{same modality} (code), and pursue better generalization by modeling them \emph{jointly}. 

We present a new approach that explicitly models the source and the target code as the same tree -- \emph{structural language modeling} (SLM).
SLM estimates the probability of the program's abstract syntax tree (AST) by decomposing it into a product of conditional probabilities over its \emph{nodes}. 
We present a neural model that computes these conditional probabilities by considering all AST paths leading to a target node, generalizing over traditional language models that consider sequences of words.
While prior work uses AST paths to \emph{read} programs \cite{alon2019code2vec}, we \emph{generate} code by predicting the next node along the set of paths, generating the target AST node-by-node. 


We evaluate SLMs on Java any-code completion, achieving a new state of the art: exact-match accuracy@1 of $18.04\%$ and accuracy@5 of $24.83\%$ (previous SOTA: $16.93\%$ and $23.17\%$).
SLMs also outperform existing models in the restricted completion task of \citet{brockschmidt2018generative} in C\# by a wide margin, $37.61\%$ accuracy@1 compared to $26.42\%$.
Our ablation study reveals the importance of \emph{joint modeling} of the source and target code, rather than separating encoders from decoders.
Finally, we discuss the theoretical advantages of SLMs, and show how they generalize many previous structural approaches for code generation.
An interactive demo of our model is presented at {\small{\url{http://AnyCodeGen.org}}}.
\section{Code Generation as Structural Language Modeling}
\label{structural}

\newcommand{\treeheight}{3.5cm} 
\newcommand{\treespace}{3mm} 
\DeclareRobustCommand\myquestionmark{\raisebox{-2pt} {\tikz[]{\node[shape=circle,draw=Maroon,fill=lightRed,inner sep=1pt]{\scriptsize{?}};}}}

\begin{figure*}[t]
\centering
\hspace{6mm}
\begin{minipage}{0.30\textwidth}
\centering
\begin{subfigure}[t]{\textwidth}
\includegraphics[height=\treeheight,keepaspectratio]{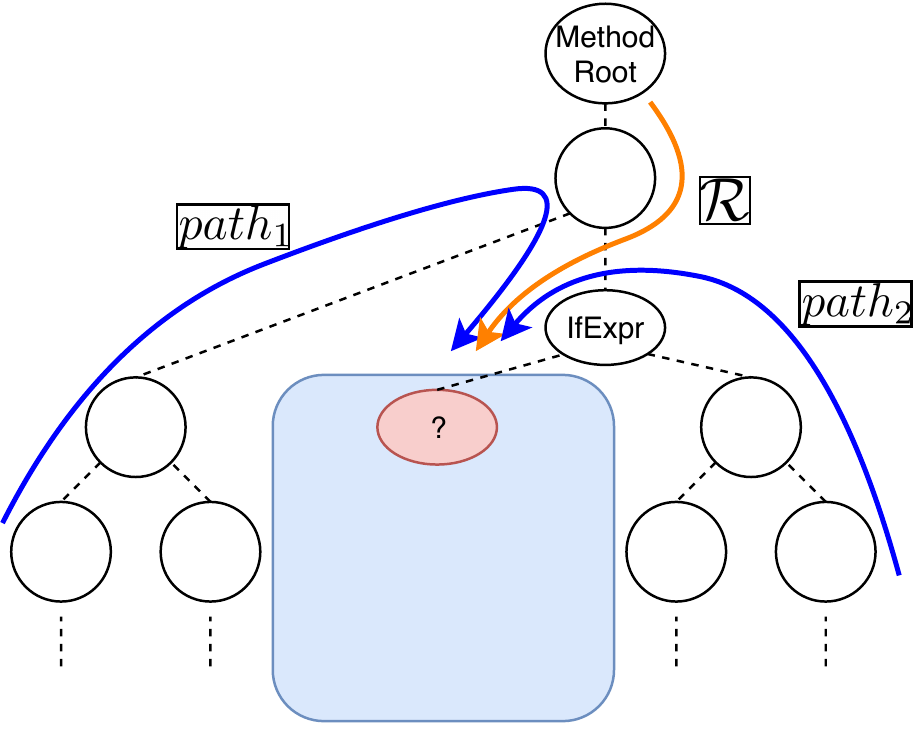}
\caption{}
\end{subfigure}
\end{minipage}
\hspace{\treespace}
\begin{minipage}{0.30\textwidth}
\centering
\begin{subfigure}[t]{\textwidth}
\includegraphics[height=\treeheight,keepaspectratio]{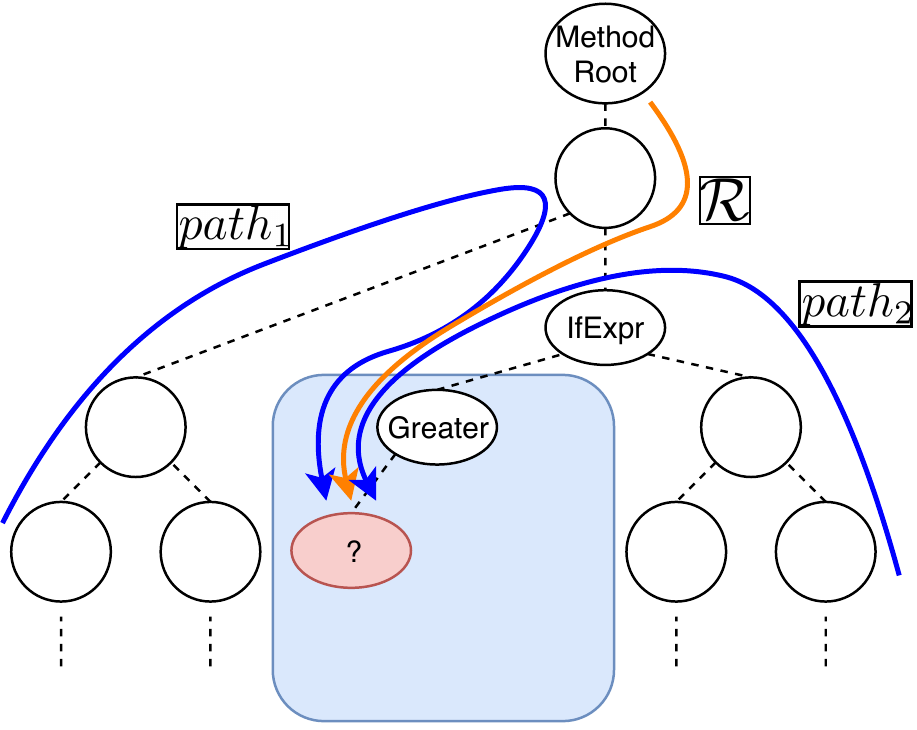}
\caption{}
\end{subfigure}
\end{minipage}
\hspace{\treespace}
\begin{minipage}{0.30\textwidth}
\centering
\begin{subfigure}[t]{\textwidth}
\includegraphics[height=\treeheight,keepaspectratio]{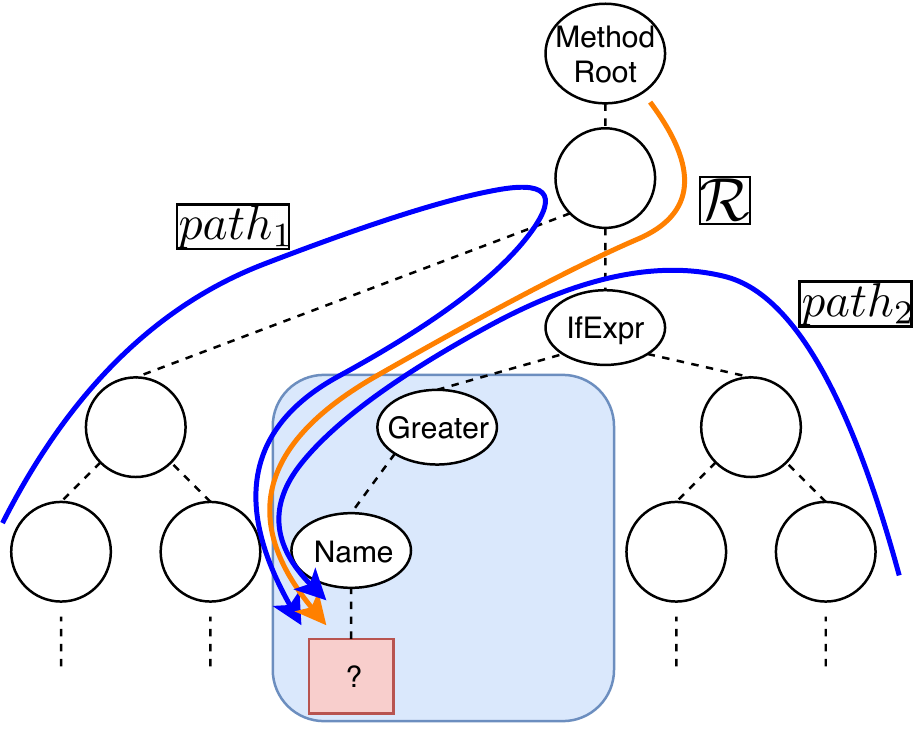}
\caption{}
\end{subfigure}
\end{minipage}
\\
\begin{minipage}{0.30\textwidth}
\centering
\begin{subfigure}[t]{\textwidth}
\includegraphics[height=\treeheight,keepaspectratio]{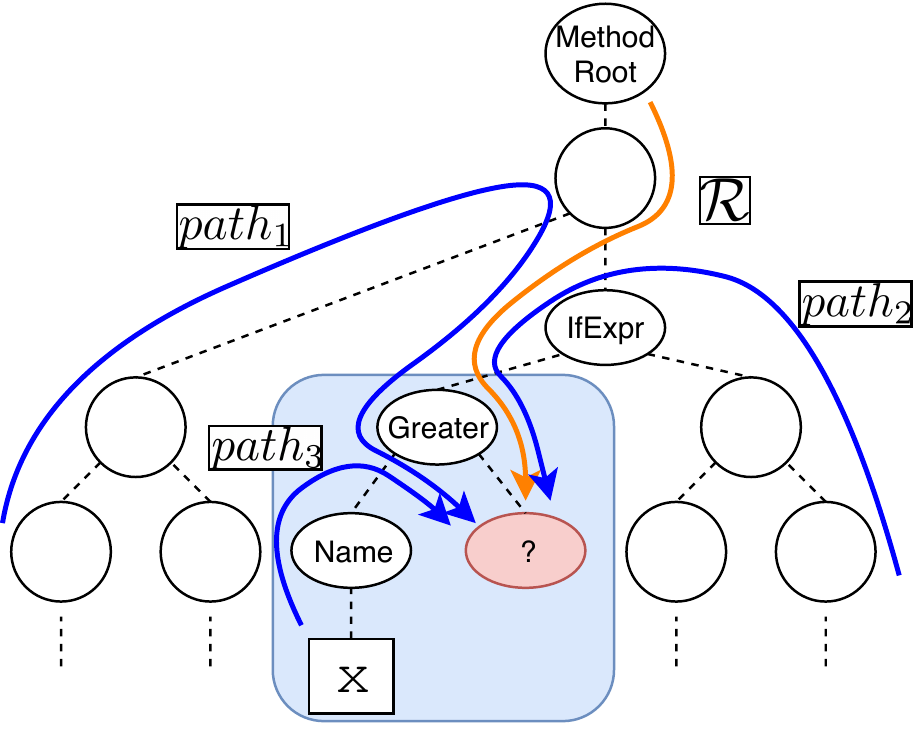}
\caption{}
\label{fig:greater:4}
\end{subfigure}
\end{minipage}
\hspace{\treespace}
\begin{minipage}{0.30\textwidth}
\centering
\begin{subfigure}[t]{\textwidth}
\includegraphics[height=\treeheight,keepaspectratio]{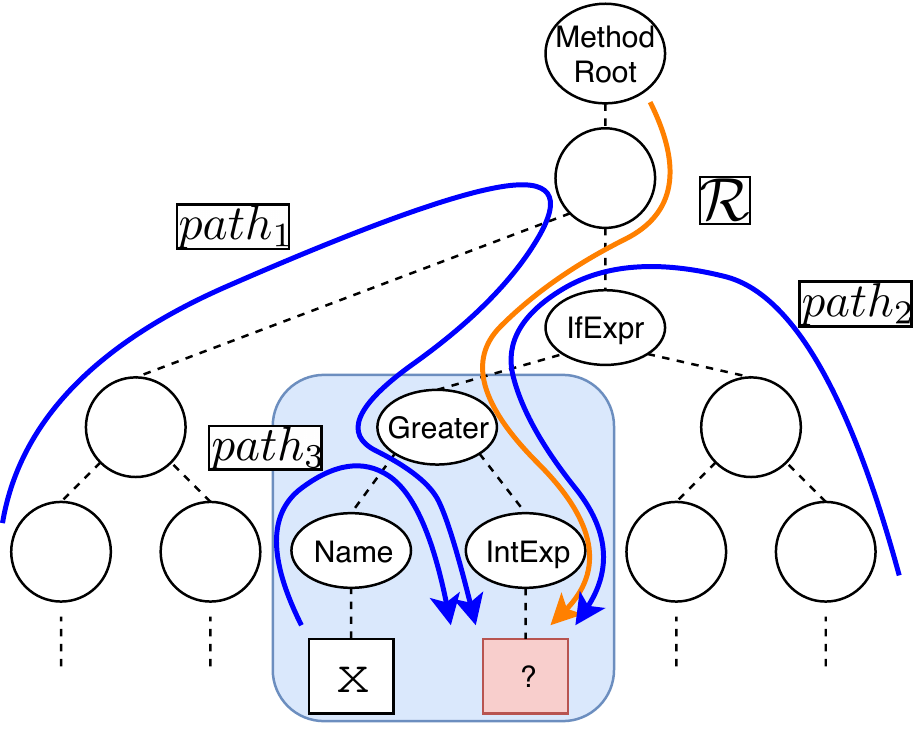}
\caption{}
\label{fig:greater:5}
\end{subfigure}
\end{minipage}
\hspace{6mm}
\begin{minipage}{0.24\textwidth}
\centering
\vspace{10mm}
\begin{subfigure}[t]{\textwidth}
\centering
\begin{minted}[fontsize=\small, frame=single,framesep=2pt,escapeinside=||]{java}
... 
if( |\setlength{\fboxrule}{1pt}\cfbox{lightblueborder}{\textbf{x > 1}}| ) { 
  ... 
} 
...
\end{minted}
\caption{}
\end{subfigure}
\end{minipage}
\caption{The subtree representing \fbox{\scode{x > 1}} is generated given its surrounding tree. At each step, the model generates the next node (denoted by \myquestionmark)
of $path_1$, $path_2$ and $path_3$ using the root path $\mathcal{R}$. Dashed lines denote the AST structure; solid lines denote AST paths. Most AST paths are omitted from the figure, for clarity.}
\label{fig:greater}
\end{figure*}
We model the task of \fulltask{} by computing the probability of a program $Pr\left(\mathcal{P}\right)$, similar to how a language model computes the probability of a natural language sentence.
While language models typically assume a \emph{sequence} as their input, our input is an abstract syntax \emph{tree} $\mathcal{A_P}$. 
We thus introduce a \emph{structural} language modeling approach (SLM). 

The intuition behind this idea is that a language model could \emph{generalize better} by modeling the \emph{tree} rather than the sequential form of the program. Further, learning from the AST allows a model to save learning capacity, instead of having to re-learn known syntactic patterns from the text. 

We first show a chain-rule decomposition of the tree's probability $Pr\left(\mathcal{A_P}\right)$ into a product of conditional \emph{node} probabilities, 
and then describe our path-based model for computing the individual conditional probabilities. 
We explain how to construct a tree from local node predictions, 
and finally discuss how our approach differs from previous work on production-based tree generation. 

\para{Representing Code as a Tree}
A program $\mathcal{P}$ is a sequence of tokens that can be unambiguously mapped to an abstract syntax tree (AST) $\mathcal{A_P}$, where every node represents an element in the language (e.g. conditions, loops, variable declarations) from a set $\mathcal{T}$.
Each AST leaf (terminal) has an associated user-defined value $v \in \mathcal{V}$.
Nonterminal nodes can have a varying number of children nodes.

\para{Decomposing the Probability of a Tree}
Given a tree $\mathcal{A_P}$, we first traverse the tree, depth-first,\footnote{Depth-first ordering is a common practice in tree generation \cite{maddison2014,raychev2016noisy}, but in principle our framework also allows for other orderings.} to induce an ordering over its nodes $a_0,\ldots,a_{|\mathcal{A_P}|} \in \mathcal{A_P}$.
We decompose the probability of a tree $Pr\left(\mathcal{A_P}\right)$ using the chain rule, akin to the standard approach in language modeling:
\begin{align}
Pr\left(\mathcal{A_P}\right) &= \prod_{t} Pr\left(a_t | a_{<t} \right)
\end{align}
where $a_{<t}$ are all the nodes that were traversed before $a_t$.

In \fulltask{}, part of the tree ($\mathcal{A_{P^{-}}}$) is already observed.
Therefore, we order the nodes of $\mathcal{A_{P^{-}}}$ to be before the nodes of the target $p$, and compute only the conditional probabilities over the nodes in $p$, essentially conditioning on the observed tree $\mathcal{A_{P^{-}}}$.

\para{Representing Partial Trees via Paths}
How can we represent the partial tree composed of $a_{<t}$ when computing $Pr\left(a_t | a_{<t} \right)$?
In standard language modeling, the structure is linear, and $a_{<t}$ is a sequence.
One way to represent a partial tree is to linearize it according to the traversal order \cite{xiao2016sequence}; however, this creates artificially long distances between the current node $a_t$ and ancestor nodes (e.g., the root $a_0$).
Another option is to use only the path from the root node to $a_t$ \cite{rabinovich2017}, but this ignores a lot of contextual information (e.g., sibling nodes). 

We follow \citet{pigeon}
and use \emph{the set of paths} from every leaf to $a_t$ together with the path from the root to $a_t$. 
Intuitively, each path captures the effect of a different, possibly distant, program element on $a_t$, along with the syntactic relationship between them. For example, in \Cref{fig:task_example} (left) the three paths originating from \scode{Path[] ret} inform the model about the existence of \scode{ret} which is an array of type \scode{Path}. Thus, when completing \scode{ret[i] = ...} -- the completion should be a \scode{Path} object. 
Other paths inform the model that the target is inside a \scode{For} loop, iterated \scode{stats.length} times. Considering the information flowing from all paths, our model correctly generates \scode{stats[i].getPath()}.

We denote the (candidate) node at time $t$ as $a_t$, its (given) parent, which is currently expanded, by $\parent$, and the set of all paths as $\mathcal{S}_t$:
\begin{align*}
\mathcal{S}_t=\left\{ \ell \leadsto \parent | \ell \in \text{leaves}\left( a_{<t} \right) \right\} 
\end{align*}
where $\ell \leadsto \parent$ is the (only) path in the tree between a leaf $\ell$ and the current node to expand $\parent$. We denote the path from the root of the program as $\mathcal{R}_t=a_0 \leadsto \parent$,
which represents the current, relative position of $\parent$ in the program (marked as $\mathcal{R}$ in \Cref{fig:greater}).
Whereas prior work used \emph{whole} paths (between two leaf nodes) to encode ASTs \cite{alon2019code2seq,alon2019code2vec}, our model observes \emph{partial} paths (between a leaf and any other node) and learns to extend them by predicting their next node.

\Cref{fig:greater} illustrates the traversal order of a subtree that represents the expression \scode{x > 1} and some of the paths used to compute the probability at each step. 
At each step, the probability of the next node is computed given the paths $\mathcal{S}_t$ from the root and every given leaf up to the current node to expand. \Cref{fig:greater:4} shows how after the terminal node with the value \scode{x} is given, $path_3$ originating from this leaf is also used to compute the probability of the next nodes.

Our path-based approach generalizes previous approaches such as ``parent feeding'' and ``previous action'' encoding \cite{yin2017}, context nodes \cite{phog16}, and some of the graph-edges of \citet{brockschmidt2018generative}.
See Section~\ref{subsec:rw_generalization} for further discussion.

\begin{wrapfigure}{R}{0.20\textwidth}
\begin{center}
\vspace{-5mm}
\input{eos_subfigure.tex}
\vspace{-5mm}
\end{wrapfigure}

\para{Generating Trees}
In sequence generation, the length of the target  sequence is controlled by generating an \scode{EOS} token to stop. When generating trees, we require a more sophisticated mechanism to control arity and depth. We augment $\mathcal{A_P}$ in two ways to allow node-by-node generation.

First, we add a special \scode{EOS}$_{node}$ node to every nonterminal to control for \emph{arity}.
Generating this node indicates that the parent node has no more children nodes.
Second, we end each subtoken sequence with a special \scode{EOS}$_{tok}$ node to control for \emph{depth} during generation; we decompose each terminal node $n_v$ into a sequence of terminal nodes $T_v$ by splitting up the node's value $v$ into \emph{subtokens} based on camel notation. 
For example, if $v=\scode{toLowerCase}$, then $T_v = \scode{to} \rightarrow \texttt{lower} \rightarrow \texttt{case} \rightarrow \scode{EOS}_{tok}$.
\Cref{fig:eos} shows an example of both \scode{EOS}$_{node}$ and \scode{EOS}$_{tok}$ in action.


\para{Node Trees vs. Production Trees}
While we predict a single \emph{node} at each step, previous work~\cite{iyer2018mapping, iyer2019learning} predicts a grammar production rule.
This representation decomposes the code in a way that often forces the model to predict with partial information.
For instance, consider generating the expression \textbf{\scode{\textcolor{Brown}{str}\textcolor{Blue}{.Substring(}\textcolor{OliveGreen}{3}\textcolor{Blue}{)}}}. 
The model of \citet{brockschmidt2018generative} would first predict the rule \textbf{\scode{Expr$\rightarrow$\textcolor{Brown}{Expr}\textcolor{Blue}{.Substring(}\textcolor{OliveGreen}{Expr}\textcolor{Blue}{)}}}, and only then expand \textbf{\textcolor{Brown}{\scode{Expr$\rightarrow$str}}} and \textbf{\textcolor{OliveGreen}{\scode{Expr$\rightarrow$3}}}.
That is, the model needs to predict the method name (\scode{Substring}) \emph{before} the invoking object (\scode{str}).
Further, the \scode{Substring} method can get either one \emph{or} two arguments, forcing the model to choose whether to use the one- or two-argument rule in advance.
Node generation, however, allows us to predict the presence of a function call and only then to predict its object and method name, rather than predicting these a priori. 



\section{Model Architecture}\label{Model}

In the previous section, we described how we can generate code given the probabilities $Pr\left(a_t|a_{<t}\right)$, where $a_{<t}$ is represented by the set of partial AST paths $\mathcal{S}_t$.
Here, we present a neural model that estimates $Pr\left(a_t | \mathcal{S}_t \right)$.
We first encode each path in $\mathcal{S}_t$ as a vector (\Cref{subsec:pathencoder});
then, we contextualize and aggregate the entire set. 
Finally, we predict the target node $a_t$ by combining a subtoken  vocabulary with a syntactic copy mechanism (\Cref{subsec:copy}).

\subsection{Encoding AST Paths}
\label{subsec:pathencoder}

Given a partial AST path, i.e., a sequence of nodes $n_1,\ldots,n_k$, our goal is to create a vector representation. 

We first represent each node $n_i$ using embeddings.
A subtoken node is represented by the index of its subtoken $w$ in the embedding matrix $E^{\mathrm{subtoken}}$;
AST nodes are represented as a pair $n_i = \left( \tau , \kappa \right)$ where $\tau$ is the node type, e.g. \scode{IfStatement}, and $\kappa$ is the node index among its sibling nodes. We represent node types using a learned embedding matrix $E^{\mathrm{type}}$ and the child indices using a learned matrix $E^{\mathrm{index}}$. The node's vector representation is the concatenation of the type and index vectors.
\begin{align*}
e \left( n_i \right) = 
\begin{cases}
E^{\mathrm{subtoken}}_w & \text{$n_i$ is the subtoken $w$} \\
\left[E^{\mathrm{type}}_{\tau};E^{\mathrm{index}}_{\kappa}\right] & \text{$n_i$ is the AST node $\left( \tau , \kappa \right)$}
\end{cases}
\end{align*}

We encode the entire path using a uni-directional LSTM stack, and take the final states:\footnote{Replacing the LSTMs with transformers yielded similar results in preliminary experiments.}
\begin{equation*}
\fpath\left(n_1,\ldots,n_k\right) = \mathrm{LSTM}\left(e\left(n_1\right),\ldots,e\left(n_k\right)\right)
\end{equation*}

Given a set of partial paths $\mathcal{S}$ (omitting the iterator $t$ for simplicity), we denote their encodings as $H\nolinebreak=\nolinebreak\{\fpath\nolinebreak\left( n_1,\ldots,n_k \right) \mid \left(n_1,\ldots,n_k\right) \in \mathcal{S}\}$.  

\para{Efficient Computation}
When modeling a subtree, there are large overlaps between paths from different time steps.
In particular, paths that originate from the same leaf share the same \emph{prefix}.
We therefore 
apply the LSTM on the prefix \emph{once} and cache the intermediate state across suffixes,
speeding up both training and inference significantly. An example is shown in \Cref{fig:efficient} (supplementary material).
\subsection{Aggregating Multiple Paths}
\label{subsec:aggregation}

Given the set of paths $\mathcal{S}$ leading up to the parent $\pi(a)$ of the target node $a$, our goal is to represent $\mathcal{S}$ in the context of predicting $a$.
To do so, we introduce the aggregation function $g\left(H,r,i \right)$. 
As its input, $g$ takes the set of encoded paths $H$, the encoded root path $r$, and the child index $i$ of the currently predicted child node $a$ relative to its parent.

We first contextualize the path encodings $H$ using a transformer encoder \cite{vaswani2017attention}.\footnote{Since $H$ is a set, we do not use positional embeddings.} In parallel, we apply a non-linear transformation to the encoding of the root path $r = \fpath \left( \mathcal{R} \right)$, in order to inform it that we wish to predict the $i$-th child of $\pi(a)$:
\begin{align*}
&Z = \mathrm{Transformer} \left( H \right) 
&\widetilde{r} = W_r \cdot \mathrm{ReLU} \left( C_{i} \cdot r \right)
\end{align*}
In this formulation, the parameter matrix $C_i$ is used when the child index is $i$, while the parameter matrix $W_r$ is used for every instance.

We then compute attention over the set of contextualized path encodings $Z$ using the index-informed root-path encoding $\widetilde{r}$ as the query; we pass the weighted average $\widetilde{z}$ and the root-path encoding $\widetilde{r}$ through another fully-connected layer; we denote the resulting vector representation as $\widetilde{h}$:
\begin{align*}
&\bm{\alpha}=\mathrm{softmax}\left(Z \cdot \widetilde{r}\right) &
& \widetilde{z} = \sum_{j} \alpha_j \cdot Z_j 
\end{align*}
\begin{equation*}
\widetilde{h} = g\left(H,r,i \right) =
\mathrm{ReLU}\left(W_{g}\left[ \widetilde{z} ; \widetilde{r} \right]\right)
\end{equation*}
where semicolons (;) denote vector concatenation.

\subsection{Predicting with a Syntactic Copy Mechanism}
\label{subsec:copy}

We can now predict $a$ from the representation $\widetilde{h}$. If the target node's parent $\pi(a)$ is a nonterminal AST node, then $a$ must be an AST node; otherwise, $a$ is a subtoken.

\para{Predicting AST Nodes}
If $a$ is an AST node, we predict $a$ using a softmax over the node type embeddings $E^\mathrm{type}$:
\begin{align*}
Pr \left( a | \mathcal{S} \right) = \mathrm{softmax} \left( E^\mathrm{type} \cdot  \widetilde{h}\right) & & \text{($\pi(a)$ is a nonterminal)}
\end{align*}
\para{Predicting Subtokens}
Programs repeatedly refer to previously declared symbols, resulting in highly repetitive usage of identifiers.
We therefore use a copy mechanism \cite{gu2016incorporating} to allow our model to predict either entire tokens or individual subtokens that exist in the context.
As we show in \Cref{sec:ablation}, copying greatly improves our model's performance. 
For brevity, we describe how entire tokens are copied, and elaborate on the copy of \emph{sub}tokens in \Cref{appendix:singlecopy}.
We score each leaf $\ell$ using a bilinear function ($W_c$) between its path's encoding $H_\ell$ and $\widetilde{h}$.
At the same time, we score the token $w$, which is the token associated with $\ell$, from a limited vocabulary using the inner product between its representation in the subtoken embedding matrix $E^\mathrm{subtoken}$ and $\widetilde{h}$.
\begin{align*}
s_\mathrm{copy} \left( \ell \right) = H_\ell \cdot W_c \cdot \widetilde{h} & & s_\mathrm{gen} \left( w \right) = E^\mathrm{subtoken}_w \cdot \widetilde{h}
\end{align*}
The scores $s_\mathrm{copy}$ and $s_\mathrm{gen}$ are then summed over all occurrences that correspond to the same symbol and subsequently
 normalized via softmax.
A key difference from most previous work \cite{ling2016latent,yin2017} is that our copy mechanism uses the \emph{syntactic} relation to the source (the path $H_\ell$), rather than the sequential relation or the graph-node representation \cite{yin2018learning}. 
\newcommand{\gnntonag}{$G\!N\!N$$\rightarrow$$\mathcal{N\!AG}$}

\section{Experimental Setup}
\label{evaluation}

\subsection{Benchmarks}
\label{subsec:benchmarks}

\para{\tasktitle{}: Java}
We take the Java-small dataset of \citet{alon2019code2seq}, which is a re-split of the dataset of \citet{conv16}. It contains 11 GitHub projects, broken down into a single method per example, and split to train/dev/test by project to reduce code overlap. This dataset was found to contain the least code duplication by \citet{allamanis2018adverse}.
We create \fulltask{} examples by selecting every expression larger than a single AST node as the target, using the remainder of the method as the context.
We remove methods containing the word ``test'' in their body or file name, and omit $10\%$ of the examples by filtering out methods longer than 20 lines to avoid configurations, initializations, and auto-generated code. 
To make the task even harder, we remove examples where the target  appears as-is in the context. 
Ultimately, this dataset contains 1.3M/10k/20k train/dev/test examples. 




\para{Restricted Completion: C\#}
To provide a fair comparison to \citet{brockschmidt2018generative}, we create an additional benchmark where the missing code is more limited. 
We use the code of \citet{brockschmidt2018generative} which filters out examples where the targets contain non-primitive types or user-defined functions. We extract the exact same types of limited expressions. 
Since the dataset of \citet{brockschmidt2018generative} is not publicly available, we consulted with Brockschmidt et al. directly and extracted examples from the raw dataset of \citet{allamanis2018learning} using their ``unseen projects test'' set. This dataset contains 30 GitHub projects broken down to one method per example. 
This dataset contains 16k/8k/3k train/dev/test examples. 

Our datasets are available at: {\small \url{http://github.com/tech-srl/slm-code-generation/}}. Detailed statistics are provided in \Cref{tab:data_stats_table} in \Cref{statistics}. 



\para{Metrics}
Following \citet{brockschmidt2018generative}, we report exact match accuracy at 1 and 5.
We also introduce a new \emph{tree@k} metric which counts a prediction as correct if the entire tree structures, ignoring leaf values, are identical.
For example, \scode{x > 1} and \scode{y > 2} would \emph{not} count as identical in \emph{exact match}, but \emph{would} count as ``tree-match identical'' because both express that an identifier is greater than an integer (\scode{NAME > INT}).
The \emph{tree@k} metric is interesting because it allows us to tease apart the model's syntactic errors from incorrect subtoken predictions.

\begin{table*}[t]
    \centering
    \small
    \begin{tabular}{lrrrr}
        \toprule
        Model          & acc@1     & acc@5 & tree@1   & tree@5  \\
        \midrule
        code2seq \cite{alon2019code2seq}  & 10.68 & 15.56 & 30.46 & 43.94  \\
        \citet{iyer2018mapping} & 5.94 & 9.19 & 25.54 & 36.75 \\
        seq2prod \cite{yin2017}  & 8.05 & 11.82 & 30.77 & 41.73  \\
        Transformer$_{\text{small}}$ \cite{vaswani2017attention}+copy &  14.23 & 21.35 & 31.83 & 47.40  \\
        Transformer$_{\text{base}}$ \cite{vaswani2017attention}+copy & 16.65 & 24.05  & 34.68 & 50.52   \\
        BiLSTM$\rightarrow$LSTM \cite{luong15}+copy & 16.93 & 23.17  & 34.29 & 49.72    \\
        seq2tree \cite{aharoni2017towards}+copy & 16.81  & 23.04  & 38.14  & 52.36  \\
        \midrule
        SLM (this work) &  \textbf{18.04} & \textbf{24.83} & \textbf{39.10} & \textbf{55.32} \\
        \bottomrule
    \end{tabular}
    \caption{Results on \fulltask{} in Java.}
    \label{tab:java-results}
\end{table*}

\subsection{Baselines}

We compare our model to a variety of original implementations and adaptations of existing models. 
We put significant effort to perform a fair comparison, including adding a copy mechanism to the NMT baselines and \emph{sub}tokenization as in our model. 
We adapt strong baselines from the literature to our task, even if they were designed to different tasks such as NL$\rightarrow$code and code$\rightarrow$NL. We re-train all the following baselines on the same datasets as our models.

\para{NMT}
We use standard autoregressive sequence-to-sequence NMT baselines, in which we subtokenize the given code snippet, replace the target in the source with a special \scode{PRED} symbol, and train the network to predict the target as a sequence of subtokens.
\emph{Transformer$_{\text{base}}$+copy} \cite{vaswani2017attention} uses the implementation of OpenNMT \cite{2017opennmt} with a copy mechanism \cite{gu2016incorporating}.
\emph{Transformer$_{\text{small}}$+copy} uses $d_{\textrm{model}}$$=$$256$, $d_{\textrm{ff}}$$=$$1024$, and 4 self attention heads per layer. \emph{BiLSTM$\rightarrow$LSTM+copy} is a 2-layer bidirectional LSTM  encoder-decoder with $d$$=$$512$ and attention. 
\emph{seq2tree+copy} follows \citet{aharoni2017towards} and learns to generate the linearized, subtokenized target AST. 

\para{Java-specific Baselines}
We use the original implementation of \citet{iyer2018mapping}, and also their \emph{seq2prod} baseline which is a re-implementation of \citet{yin2017};
these are designed for NL$\rightarrow$code tasks, in which we feed the code context as the NL input.
The model of \citet{iyer2018mapping} is designed to get additional input of the available variables \emph{and their types}, for which we do not feed types.
While these models could also be applied to other languages, their implementation only supports Java.

\para{C\#-specific Baselines}
We compare our model to the graph-based \gnntonag{} model using the implementation of \citet{brockschmidt2018generative}. 
\citet{phog16} kindly trained and tested their non-neural PHOG model on our C\# dataset. 
We note that PHOG does not have an explicit copy mechanism, and considers only context to the left of the target code, while we consider also context to the right. 
Extending PHOG 
could potentially improve its results.

In both Java and C\#, we compare to \emph{code2seq} \cite{alon2019code2seq}, which is a strong code$\rightarrow$NL model. We train it to generate the target code as a \emph{sequence} of subtokens.



\subsection{Implementation and Hyperparameter Settings}

\para{Architecture}
We use embeddings of size $512$, $2$ layers of LSTMs with $256$ units, and $4$ transformer layers with 8 attention heads. We kept a small subtoken vocabulary of size $1000$ to encourage the model to learn to copy; larger vocabularies did not show an improvement. 
These resulted in a very lightweight model of only 15M parameters, which is close to \emph{Transformer$_{\text{small}}$} (11.8M parameters). In comparison, \emph{Transformer$_{\text{base}}$} had more than 45M parameters ($3\times$ more parameters than our model). 

\para{Training} We train the model end-to-end on a single V100 GPU, using cross entropy and the Adam optimizer \cite{kingma2014adam}, an initial learning rate of $10^{-4}$ multiplied by $0.95$ every $20k$ steps. We bucket examples based on the number of predictions in the target subtree (nodes + subtokens + \scode{EOS}), and vary the batch size such that each batch contains about $512$ targets. We train the model to prefer copying entire tokens rather than copying subtokens, if possible, by optimizing for the entire token as the true label. We apply dropout of $0.25$ in the Transformer layers, and a recurrent dropout of $0.5$ in the LSTMs.

\para{Inference}
We perform beam search with width of $5$ and optimize for accuracy@1.

\begin{table}[h!]
    \small
    \begin{minipage}{0.49\textwidth}
        \begin{tabular}{lrrrr}
            \toprule
            Model & acc@1     & acc@5 & tree@1   & tree@5 \\
            \midrule
            $G\!N\!N\!\!\rightarrow\!\!\mathcal{N\!\!AG}$  & 15.19 & 27.05 & 26.48 & 40.09 \\
            code2seq & 6.20 & 10.05 & 21.97 & 30.89 \\
            seq2seq+copy & 26.42 & 37.94 & 34.10 & 49.23 \\ 
            seq2tree+copy & 22.29  & 35.86 & 31.85 & 48.53 \\
            PHOG  & 7.40 & 12.00 & -- & -- \\
            \midrule
            SLM (this work) & \textbf{37.61} & \textbf{45.51} & \textbf{51.10} & \textbf{59.82} \\
            \bottomrule
        \end{tabular}
        \caption{Results on \restrictgen{} in C\#.}
        \label{tab:csharp-limited-results}
    \end{minipage}
    
\end{table}

\section{Results}	
\label{sec:results}

\para{\tasktitle{}: Java} 
\Cref{tab:java-results} shows that our SLM achieves over $1.1\%$ and $0.78\%$ better \emph{acc@1} and \emph{acc@5} (respectively) over the two strongest baselines.
The improvement over \emph{Transformer$_{\text{small}}$}, which is closer to our model in the number of parameters, is even higher: over 3.8\% and 3.4\% in \emph{acc@1} and \emph{acc@5}.

The NMT baselines performed better than code-specific baselines. 
We hypothesize that the reason is that the NMT baselines are more generic, while the code-specific baselines are designed for different tasks:
\emph{seq2prod} is designed for tasks which involve generating code \emph{given natural language input};
\citet{iyer2018mapping} additionally expects all member methods, fields, and their types as input;
\emph{code2seq} is designed to generate sequences rather than code, and does not have a copy mechanism.
An approximation of \emph{code2seq} with a copy mechanism is presented in \Cref{sec:ablation}. 

Interestingly, the syntactically-informed \emph{seq2tree} baseline achieved the highest \emph{tree@k} among the baselines, while our model achieved higher \emph{acc@k} and \emph{tree@k}. This shows that leveraging the syntax can benefit NMT models as well.

\para{Restricted Completion: C\#}
\Cref{tab:csharp-limited-results} shows the results for the \restrictgen{} task in C\#, where \emph{seq2seq+copy} is the \emph{BiLSTM$\rightarrow$LSTM+copy} model which performed the best among the Java baselines.
We first observe that the \emph{seq2seq+copy} and the \emph{seq2tree+copy} baselines outperform the \gnntonag{} of \citet{brockschmidt2018generative}, who introduced this task.
Although \citet{brockschmidt2018generative} did compare to a seq2seq baseline, their \gnntonag{} model could copy symbols from the context, but their baseline did not.
To conduct a fair comparison with our SLM model, we equipped the seq2seq and seq2tree baselines with a copy mechanism. 
Even though the \emph{seq2seq+copy} and the \emph{seq2tree+copy} baselines perform substantially better than the state of the art in this setting, our SLM model is able to go beyond, achieving significant gains over all models.

The superiority of our model over \gnntonag{} may also be related to the GNN bottleneck \cite{alon2020bottleneck}, which hinders GNNs from propagating long-range messages. In contrast, propagating long-range messages using paths is natural for our model.

\begin{table}[t]
\begin{minipage}{0.48\textwidth}
        \begin{tabular}{lrrrr}
            \toprule
            Ablation   & acc@1     & acc@5 & tree@1   & tree@5 \\
            \midrule
            Paths$\rightarrow$Seq & 12.95 & 18.52 & 33.44 & 43.43 \\
            Seq$\rightarrow$Path & 12.12 & 17.12 & 28.68 & 43.99 \\
            Paths$\rightarrow$Paths & 17.63 & 24.62 & 37.78 & 53.98\\
            No Root Att & 14.43 & 18.48 & 28.20  & 35.65\\
            No Copy & 10.72 & 15.70 & 30.61 & 44.35\\
            \midrule
            SLM (original) &  18.04 & 24.83 & 39.10 & 55.32 \\
            \bottomrule
        \end{tabular}
        \caption{Ablations on \fulltask{} in Java.}
        \label{tab:ablation-results}
    \end{minipage}
\end{table}
    
\section{Ablation Study}
\label{sec:ablation}
To understand the importance of the various components and design decisions in our model, we conducted an extensive ablation study.

\textbf{\emph{Paths$\rightarrow$Seq}} follows \emph{code2seq} \cite{alon2019code2seq} and separates the model to an encoder and a decoder, where the decoder generates the target code as a sequence of subtokens. The main difference from \emph{code2seq} is that \emph{Paths$\rightarrow$Seq} includes a copy mechanism, as in our model.

\textbf{\emph{Seq$\rightarrow$Path}} follows \citet{rabinovich2017} and separates our model to an encoder and a decoder (including a copy mechanism), where the encoder encodes the context as a sequence of subtokens using a BiLSTM, and the decoder generates the missing subtree using the root path and the index of the generated child. 

\textbf{\emph{Paths$\rightarrow$Paths}} is similar to our SLM model except that it uses separate encoder and decoder. 
These encoder and decoder have untied weights, unlike our SLM model which models the source and the target jointly.


\textbf{\emph{No Root Attention}} uses max pooling instead of attention in aggregating multiple paths (see Section~\ref{subsec:aggregation}). The index-informed path from the root to the target's parent ($\mathcal{R}$ in \Cref{fig:greater}) is concatenated with the result, instead of being used as attention query.

\textbf{\emph{No Copy}} replaces copy mechanism with a much larger vocabulary (25k subtokens instead of 1k).


\para{Results} \Cref{tab:ablation-results} shows the results of these alternatives. 
As our SLM model performs better than \emph{Paths$\rightarrow$Paths}, this ablation shows the importance of joint modeling of the context and the target subtree by parameter tying.

Each of \emph{Paths$\rightarrow$Paths} and the seq2seq baselines (\Cref{tab:java-results}) performs better than \emph{Paths$\rightarrow$Seq} and \emph{Seq$\rightarrow$Path}; this shows the importance of \emph{using the same type of encoder and decoder} for \fulltask, rather than combining ``an optimal encoder'' with ``an optimal decoder''. 
While this distinction between encoder and decoder types might be necessary for semantic parsing \cite{rabinovich2017, dong2018coarse}, NL$\rightarrow$code \cite{yin2017} and code$\rightarrow$NL \cite{alon2019code2seq, fernandes2018structured} tasks because of the different modalities of the input and the output, this discrepancy may hurt generalization when the output is essentially a missing part of the input's AST. 

\emph{Paths$\rightarrow$Paths} performs better than the seq2seq baselines (\Cref{tab:java-results}), showing the advantage of using paths over textual sequences, even without parameter tying.





\emph{No Root Attention} degrades \emph{acc@1} and \emph{acc@5} by 3.6\% to 6.3\%. 
This shows that
dynamically attending to the context paths given the current root path is crucial. 

\emph{Not using a copying mechanism} results in a degradation of 7.3\% to 9.1\%. Programs use symbols and identifiers repetitively, thus the ability to copy symbols from the context is crucial for this task.
For this reason, we included a copying mechanism in all NMT baselines in \Cref{evaluation}.


\newcommand{\gear}[1]{%
\raisebox{-1pt}{
  \begingroup\normalfont
  \begin{tikzpicture}
\def\teethnumber{7}     
\def\threadheight{0.8pt}   
\def\outrad{2.3pt}
\draw[fill=gray,even odd rule] 
let 
\n{dpt} = {360/\teethnumber)}
in
(0,0) circle (1.1pt)                          
({\n{dpt}*(0.5)}:\outrad) \foreach \x in {1,...,\teethnumber}{ 
arc ({\n{dpt}*(\x-0.5)}:{\n{dpt}*(\x-0.25)}:\outrad)  
--++(\x*\n{dpt}:\threadheight)                        
arc ({\n{dpt}*(\x-0.25)}:{\n{dpt}*(\x+0.25)}:\outrad) 
--++(\x*\n{dpt}:-\threadheight)                       
arc ({\n{dpt}*(\x+0.25)}:{\n{dpt}*(\x+0.5)}:\outrad)--
({\n{dpt}*(\x+0.5)}:\outrad)
};
\end{tikzpicture}
  \endgroup}
}

\makeatletter
\newcommand\RSloop{\@ifnextchar\bgroup\RSloopa\RSloopb}
\makeatother
\newcommand\RSloopa[1]{\bgroup\RSloop#1\relax\egroup\RSloop}
\newcommand\RSloopb[1]%
  {\ifx\relax#1%
   \else
     \ifcsname RS:#1\endcsname
       \csname RS:#1\endcsname
     \else
       \GenericError{(RS)}{RS Error: operator #1 undefined}{}{}%
     \fi
   \expandafter\RSloop
   \fi
  }
\newcommand\X{0}
\newcommand\RS[1]%
  {\begin{tikzpicture}
     [every node/.style=
       {circle,draw,fill,minimum size=1.5pt,inner sep=0pt,outer sep=0pt},
      line cap=round
     ]
   \coordinate(\X) at (0,0);
   \RSloop{#1}\relax
   \end{tikzpicture}
  }
\newcommand\RSdef[1]{\expandafter\def\csname RS:#1\endcsname}
\newlength\RSu
\RSu=1ex
\RSdef{i}{\draw (\X) -- +(-90:\RSu) node{};}
\RSdef{l}{\draw (\X) -- +(-135:\RSu) node{};}
\RSdef{r}{\draw (\X) -- +(-45:\RSu) node{};}
\RSdef{I}{\draw (\X) -- +(90:\RSu) coordinate(\X I);\edef\X{\X I}}
\RSdef{L}{\draw (\X) -- +(135:\RSu) coordinate(\X L);\edef\X{\X L}}
\RSdef{R}{\draw (\X) -- +(45:\RSu) coordinate(\X R);\edef\X{\X R}}

\newcommand\mytree{\RS{lir}}

\begin{figure*}[h!]
\begin{minipage}{0.48\textwidth}
\raggedright
\begin{subfigure}{0.995\textwidth}
\begin{minted}[fontsize=\footnotesize, frame=single,framesep=2pt,escapeinside=||]{Java}
private static void log(String value) {
  if (value != null 
      && |\greenbox{\phantom{value.length() > 55}}|)
    value = value.substring(0, 55)+"...";
  LOG.info(value);
}


\end{minted}
\footnotesize
\centering
\begin{tabular}{ll}
\toprule
\multicolumn{2}{l}{True ref:                \qquad\quad\ \ \ \scode{value.length() > 55}} \\
\midrule
\multirow{5}{*}{\makecell[l]{SLM\\top-5:}} & (9.6\%)\quad  \scode{value.length() > 0} \quad\quad\quad \mytree \\
 & (7.3\%)\quad \textbf{\scode{value.length() > 55}} \quad\quad\, \checkmark \\
 & (1.8\%)\quad   \scode{value.startsWith("...")} \\
 & (1.5\%)\quad  \scode{!value.startsWith("...")} \\
 & (0.9\%)\quad  \scode{value.charAt(0) == \textquotesingle.\textquotesingle}\\
\bottomrule
\end{tabular}
\caption{}
\label{fig:value-length-figure}
\end{subfigure}
\end{minipage}
\begin{minipage}{0.51\textwidth}
\raggedleft
\begin{subfigure}{0.98\textwidth}
\begin{minted}[fontsize=\footnotesize, frame=single,framesep=2pt,escapeinside=||]{Java}
public int compareTo(LongWritable o) {
    long thisValue = this.value;
    long thatValue = o.value;
    return (thisValue < thatValue ? -1 : 
        (|\greenbox{\phantom{thisValue == thatValue ? 0 : 1}}|));
}
\end{minted}
\footnotesize
\begin{tabular}{ll}
\toprule
\qquad\quad\ \ \ \scode{thisValue == thatValue ? 0 : 1} \\
\midrule
(16.3\%)\quad  \scode{thisValue == thisValue ? 0 : 1}  & \mytree  \\
                                (11.0\%)\quad  \scode{\textbf{thisValue == thatValue ? 0 : 1}} & \checkmark\\
                                \;\;(9.5\%)\quad  \scode{thisValue == value ? 0 : 1} &  \mytree \\
                                \multicolumn{2}{l}{\;\;(6.6\%)\quad  \scode{thisValue > thatValue ? 0 : 1}}  \\
                                \;\;(6.1\%)\quad  \scode{(thisValue == thatValue)} \scode{?} \scode{0 : 1} & $\leftrightarrow$  \\

\bottomrule
\end{tabular}
\caption{}
\label{fig:taskid-figure}
\end{subfigure}
\end{minipage}
\caption{Examples for cases where the top candidate is a ``tree-match'' (marked with \mytree), but only the second candidate is an ``exact match'' (marked with \checkmark{} in bold). Predictions that are logically equivalent to the ground truth are marked with $\leftrightarrow$. Additional (and larger) examples along with the predictions of the baselines are shown in \Cref{appendix:java_examples,appendix:csharp_examples}.}
\label{fig:qualitative}
\end{figure*}

\section{Qualitative Analysis}\label{sec:quali}
\begin{figure*}[t]
\centering
\begin{minipage}{0.7\textwidth}
\begin{minted}[fontsize=\footnotesize, frame=single,framesep=2pt,escapeinside=||]{Java}
public float getProgress() {
    this.readLock.lock();
    try {
        if (this.currentAttempt != null) {
            return |\greenbox{\phantom{this.currentAttempt.getProgress()}}|;
        }
        return 0;
    } finally {
      this.readLock.unlock();
    }
}
\end{minted}
\end{minipage}
\vspace{1mm}
\footnotesize
\centering
\begin{tabular}{lrlccc}
\toprule
True ref:                            & &  \multicolumn{3}{l}{\texttt{this.currentAttempt.getProgress()}} & \\
\midrule
\multirow{5}{*}{SLM top-5:}             & (31.3\%) & \texttt{this.currentAttempt.getCount()}   & & \mytree & \\
                                 & (30.6\%) & \texttt{-1}  & & & \gear{} \\
                                 & (1.5\%) & \texttt{this.currentAttempt.get()}  & & \mytree & \\
                                 & (1.2\%) & \texttt{this.currentAttempt.getTime()}  & & \mytree & \\
                                 & (0.9\%) & \texttt{\textbf{this.currentAttempt.getProgress()}}  & \checkmark & \mytree & \gear{}  \\ 
\bottomrule
\end{tabular}

\caption{An example from our test set in which a compiler-guided generation could filter out non-compiling candidates, and thus rank the ground truth \emph{second} instead of \emph{fifth}. Four out of the five candidates are ``tree-match'' (marked with \mytree), the fifth candidate is an ``exact match'' (marked with \checkmark{} in bold), and only the second and the fifth candidate compile (marked with \gear{}).}
\label{compilation_check_figure}
\end{figure*}
Our main results (\Cref{tab:java-results} and \Cref{tab:csharp-limited-results}) reveal a gap between \emph{acc@k} and \emph{tree@k}: when ignoring identifier values and comparing only the tree structure, accuracy is significantly higher across all models. While our SLM model performs better than all baselines in \emph{acc@k}, our model also shows greater potential for improvement in its \emph{tree@k} results, which are much higher than the baselines'.
We thus focus on studying the cases where the tree was predicted correctly, but the model failed to generate the code exactly including names.


\Cref{fig:value-length-figure} shows an example of this case: the ground truth has a structure of the form: \scode{NAME.NAME() > INT}. Our model predicts \scode{value.length() > 0} (a tree-match) as its first candidate and \scode{value.length() > 55} (the ground truth) as its second. Null-checking a string is often followed by checking that it is also not empty, making the first candidate a reasonable prediction as well.

\Cref{fig:taskid-figure} shows another example: in this case, the ground truth \scode{thisValue == thatValue ? 0 : 1} was predicted correctly only as the second candidate. Nevertheless, the top-3 candidates are tree-matches since all of them are of the form: \scode{NAME == NAME ? INT : INT}. Interestingly, the fifth candidate \scode{(thisValue == thatValue) ? 0 : 1} is logically-equivalent to the ground truth. 

In both examples, our model's top candidate differs from the ground truth by \emph{a single identifier or literal}: in \Cref{fig:value-length-figure} the model predicted \scode{0} instead of \scode{55}; in \Cref{fig:taskid-figure} the model predicted \scode{thisValue} instead of \scode{thatValue}. Such single \emph{sub}token errors are responsible for $30\%$ of the cases where the model's top prediction is a tree-match but not an exact match. Single \emph{token} (whole identifier or literal) mismatches are responsible for $74\%$ of these cases. Thus, improving our model's ability to predict the right names has the potential to enhance our gains furthermore. Detailed results of allowing such mistakes in our model and in the baselines can be found in \Cref{quali-supp}. 

Additional possible post-filtering could filter out candidates that do not compile. In \Cref{compilation_check_figure}, the first, third and fourth candidates do not compile, because the \scode{this.currentAttempt} object does not have \scode{getCount}, \scode{get}, nor \scode{getTime} methods. If the model's predictions would have been considered in the context of the entire project including its dependencies, these candidates could have been filtered out, and the (correct) fifth candidate would be ranked \emph{second}. We leave compiler-guided code generation to future work.

Additional examples can be found in \Cref{appendix:java_examples,appendix:csharp_examples},  and in our interactive demo at {\small \url{http://AnyCodeGen.org}}.


\section{Related Work}
\label{subsec:rw_generalization}

\para{Generalizing Previous Approaches} Our approach frames code generation as predicting the next node in all partial AST paths. This simple framing generalizes most previous work, without hand-crafted edges and special actions:
\begin{itemize}[itemsep=0pt,nosep, leftmargin=20pt]
	\item Models that use information about ancestor nodes only \cite{rabinovich2017}, as well as the ``Parent Feeding'' of \citet{yin2017}, are generalized by our model, since all paths that go into a node $a_t$ pass through its parent, and the path from the root 
	is the attention query.
	\item The ``previous action encoding'' of \citet{yin2017} is also a special case of our approach, because $\mathcal{S}_t$ contains the paths starting from the \emph{previously expanded} leaves of $\mathcal{A}_p$ into the currently expanded node $\parent$, such as $path_3$ in \Cref{fig:greater:5}. 
	\item The ``context node'' of PHOG \cite{phog16} is just one of the previously-traversed leaf nodes in $a_{<t}$. 
Thus, not only that our model conditions on this context node as well, our model also takes into account the \emph{syntactic relation}, i.e., the path, between the context and $\parent$. 
Moreover, while PHOG conditions on a single leaf, SLMs condition on \emph{every} leaf in $a_{<t}$.
	\item Finally, \citet{brockschmidt2018generative} define special graph edges (e.g., ``NextSib'' and ``Child'') to capture relations on the AST.
\citet{allamanis2018learning} further defines data-flow and control-flow graph edges such as ``ComputedFrom'' and ``GuardedByNegation''.
Most of these relations can be expressed as partial AST paths without manually designing them.
\end{itemize} 
\para{Program Generation} Learning to generate programs is one of the oldest problems in machine learning \cite{waldinger1969prow} and has been considered by some as the ``holy grail of computer science'' \cite{pnueli1989synthesis, gulwani2017program}. Typically, the task is to generate a program given some form of input or context, such as complete formal specifications \cite{green1981application,si2018learning} or input-output examples \cite{gulwani2011automating, devlin2017robustfill,parisotto2016neuro,balog2016deepcoder,gaunt2017differentiable}.
While these approaches work well in some cases, they are often bounded to DSLs that prevent them from being applied to realistic, general-purpose code. 

\citet{phog16} learn a dynamic DSL expression that points to a \emph{single} context that guides the generation of a JavaScript program.
\citet{maddison2014} and \citet{amodio2017neural} generate general-purpose unconditional code, 
and do not deal with the challenge of fitting the code to a given context.  

\citet{brockschmidt2018generative} addressed a similar code completion task as ours using a graph encoder and a neural attribute grammar decoder. 
However, they limit their model to generate only 
primitive types or arrays of these; use a closed vocabulary; and omit 
user-defined functions. 
In this paper, we lift these constraints and allow any, general-purpose, generation of code, of all types and containing any names. 
As we show in  \Cref{sec:results}, our model performs significantly better. 

\citet{murali2018bayou} generate code given a set of APIs in a "Java-like" language; they state that their approach is thus intrinsically limited to generate only API-heavy programs.
\citet{yin2018learning} generate general-purpose code by applying a given edit to a given code snippet. 
\citet{brody2020neural} predict code edits directly given other edits that occurred in the context.
\citet{yin2017} and \citet{rabinovich2017} used a top-down syntactic approach for generating general-purpose code given a natural language description. Models that address APIs$\rightarrow$code, edit$\rightarrow$code, or NL$\rightarrow$code tasks must model the input separately and differently from the output code. 
As we show in \Cref{sec:ablation}, modeling the source and the target differently perform poorly in our task, in which the input is code as well.



\citet{chen2018tree} addressed JavaScript$\leftrightarrow$CoffeeScript translation with a tree-to-tree approach, which required a strong alignment between the source and target trees. 


\section{Conclusion}\label{Conclusion}
We presented a novel approach for \fulltask{}: joint modeling of an AST and its missing subtree using a structural language model. 
Our approach generalizes most previous work in this area while reaching state-of-the-art performance on challenging benchmarks.
We demonstrate our approach in generating general-purpose code, in restricted and unrestricted settings, in two languages. Our model outperforms  a variety of strong baselines, including programming language-oriented models and strong NMT models applied in our settings.

We believe that structural language modeling enables a wide range of future applications, similarly to how language modeling research has contributed to NLP in recent years.
Our approach also has a variety of direct applications such as code completion, detecting and fixing unlikely existing code, and re-ranking solutions produced by another synthesizer or solver. 
To these ends, 
we make all our code, datasets, and trained models publicly available.

\section*{Acknowledgments}
We would like to thank Guy Waldman for developing the \href{http://AnyCodeGen.org}{AnyCodeGen.org} website,
Pavol Bielik and Martin Vechev for training their PHOG model on our dataset,
Srinivasan Iyer for his useful advice, the guidance in training his model and adapting it to our task,
Marc Brockschmidt for his useful implementation tips and guidance in training his model, 
and Miltiadis Allamanis for the guidance in reproducing his C\# dataset.


\onecolumn
\section*{Supplementary Material}
\appendix
\begin{figure*}[!h]
\begin{minipage}{0.5\textwidth}
\small
\centering
\begin{tabular}{lrr}
\toprule
                                & Java    & C\# \\
\midrule
\#projects - training           & 9             &  25                \\
\#projects - validation         & 1             & 2  \\
\#projects - test               & 1             & 3  \\
\#examples - training           & 1,309,842       & 16,295     \\
\#examples - validation         & 10,000        & 8,183         \\
\#examples - test               & 20,000        &  3,305        \\
Avg. number of paths  			& 27.8           &  131.1            \\
Avg. source length - lines  		& 10.4 		& 57.5   \\
Avg. source length - tokens     & 77.7            & 264.3                 \\
Avg. source length - subtokens  & 100.6            & 343.6                \\
Avg. target length - tokens     & 5.4             & 3.9            \\
Avg. target length - subtokens  & 7.8             &  5.0           \\
Avg. target length - tree nodes & 3.8 		      & 3.9 \\
Avg. target length - tree targets & 10.8 & 10.8    \\
\bottomrule
\end{tabular}
\caption{Statistics of our datasets. When not mentioned otherwise, the statistic was measured on the training set.}
\label{tab:data_stats_table}
\end{minipage}
\hspace{8mm}
\begin{minipage}{0.4\textwidth}
	\input{efficient_figure.tex}
\end{minipage}
\end{figure*}
\section{Data Statistics}\label{statistics}
\Cref{tab:data_stats_table} shows some statistics of our used datasets. In Java: for the validation set, we randomly sampled $10,000$ examples from the raw validation set; for the test set, we randomly sampled $20,000$ examples from the raw test set. 

We will release all datasets, raw and preprocessed, with the final version.

\section{Additional Evaluation Details}\label{evaluation-details}
For both Java and C\# models, we experimented with the following hyper-parameter values. We performed beam search on the validation set after every training iteration, and we selected the best configuration and checkpoint according to accuracy@1 on the validation set. After the best configuration was chosen, we ran a single evaluation run on the test set. 
\begin{itemize}
	\item $\fpath \in \{LSTM,Transformer\}$ -- how to encode each path.
	\item LSTM \#layers $\in \{1,2\}$
	\item $d_{subtoken} \in \{256,512\}$ -- embedding size.
	\item Transformer layers $\in \{0,1,2,3,4\}$
	\item $lr \in \{10^{-3},10^{-4},10^{-5}\}$ -- learning rate
	\item Learning rate decay every $\{10000, 20000, 40000\}$ steps.
\end{itemize}

\section{Qualitative Analysis cont. - Correct Tree, Incorrect Names}\label{quali-supp}
\begin{table}[!h]
    \centering
    \small
    \begin{tabular}{lrrrrrrrrrrr}
        \toprule
        \multirow{2}{*}{Model}  & \multicolumn{2}{c}{Exact-match (acc@k)}  & & \multicolumn{2}{c}{One SubToken Diff} &  & \multicolumn{2}{c}{One Token Diff} &  & \multicolumn{2}{c}{Tree@k}  \\
        \cline{2-3} \cline{5-6} \cline{8-9} \cline{11-12} 
          & @1 & @5 &  & @1 & @5 &  & @1 & @5 &  & @1 & @5 \\
        \midrule
        Transformer$_{\text{base}}$ +copy & 16.65 & 24.05  &  & 23.08 & 34.06 &  &29.39 & 43.46 &  & 34.68 & 50.52   \\
        BiLSTM$\rightarrow$LSTM +copy & 16.93 & 23.17  &  &  22.39& 31.68 &  & 27.23 &38.92  & & 34.29 & 49.72    \\
        seq2tree +copy & 16.81  & 23.04   & & 24.02 & 33.89 &  & 32.67 & 43.75 &  & 38.14  & 52.36  \\
        \midrule
        SLM (this work) & \textbf{18.04} & \textbf{24.83} &  & \textbf{24.40} & \textbf{35.19} &  & \textbf{33.68} & \textbf{46.57} &  &  \textbf{39.10} & \textbf{55.32} \\
        \bottomrule
    \end{tabular}
    \caption{Examining the gap between \emph{acc@k} and \emph{tree@k}: the \emph{acc@k} and \emph{tree@k} results here are the same as in \Cref{tab:java-results}; \emph{One SubToken Diff} allows a single \emph{sub}token mismatch; \emph{One Token Diff} allows a single token mismatch. }
    \label{tab:one-diff}
\end{table} 
In \Cref{sec:quali} we discussed the gap between \emph{acc@k} and \emph{tree@k}. We found that 30\% of the examples in the gap could have been \emph{exact match} if a single subtoken prediction was fixed; 74\% of the examples in the gap could have been \emph{exact match} if a single identifier prediction was fixed. \Cref{tab:one-diff} shows the accuracy of our model and the leading baselines if a single subtoken or a single token mismatches were counted as correct: \emph{One SubToken Diff} and \emph{One Token Diff} are similar to \emph{exact match}, except that they allow a single subtoken or a single token mistake, respectively. As \Cref{tab:one-diff} shows, not only that our model performs better than the baselines in \emph{exact match}, it also shows a greater potential for improvement.

\section{Copying Single Subtokens}
\label{appendix:singlecopy}
In addition to scoring the entire token to be copied, we also score each of the subtokens composing it according to their position. For each position $i$, we add a scoring function $s_{copy_i}$, such that $s_{copy_i}\left(\ell\right)$ produces the copying score of the $i$'th subtoken of $\ell$, which we denote as $\ell_i$:

\begin{align*}
s_w = s_\mathrm{gen}\left(w\right) 
+ \sum_{\mathrm{val}\left(\ell\right) = w} s_\mathrm{copy\_token}\left(\ell\right)
+ \sum_{i}\sum_{\mathrm{val}\left(\ell_i\right) = w} s_\mathrm{copy_i}\left(\ell\right)
\end{align*}
\begin{align*}
Pr\left(a | \mathcal{S} \right) = \mathrm{softmax} \left( s \right)
\end{align*}

Where $s_\mathrm{copy\_token}$ is the scoring function of copying the entire token, described in \Cref{subsec:copy}.

For example, a token of \scode{getX} is scored entirely using $s_\mathrm{copy\_token}$; each of its subtokens, \scode{get} and \scode{X}, are scored using $s_{copy_\mathbf{1}}$ and $s_{copy_\mathbf{2}}$ respectively. 
That is, the model can either copy the entire token, or copy only some of its subtokens.
This  ability is especially useful in generating a name like \scode{setX}, where \scode{getX} appears in the context, and \scode{X} is any unknown, user-defined, subtoken; the model learns to generate \scode{set} from the vocabulary, and copy only the subtoken \scode{X}.

\definecolor{darkblue}{HTML}{000080}

\begin{figure*}[h!]
\centering
\begin{subfigure}[t!]{1.0\textwidth}
\begin{minted}[fontsize=\footnotesize, frame=single,framesep=2pt,escapeinside=**]{Java}
protected void checkRpcAdminAccess() throws IOException, AccessControlException {
  UserGroupInformation ugi = UserGroupInformation.getCurrentUser();
  UserGroupInformation zkfcUgi = UserGroupInformation.getLoginUser();
  if (adminAcl.isUserAllowed(ugi)
    || ugi.getShortUserName().equals(*\greenbox{zkfcUgi.getShortUserName()}*)) {
      LOG.info("Allowed RPC access from " + ugi 
        + " at " + Server.getRemoteAddress());
      return;
    }
  String msg = "Disallowed RPC access from " + ugi 
    + " at " + Server.getRemoteAddress() + ". Not listed in " + DFSConfigKeys.DFS_ADMIN;
  LOG.warn(msg);
  throw new AccessControlException(msg);
}
\end{minted}
\end{subfigure}

\footnotesize
\centering
\begin{subfigure}[t!]{1.0\textwidth}
\centering
\begin{tabular}{llrl}
\toprule
True ref:                            & \texttt{zkfcUgi.getShortUserName()} & &\\
\midrule
\multirow{3}{*}{SLM top-5 candidates:}             & \textbf{\texttt{{\color{Brown}zkfcUgi}.{\color{Brown}getShortUserName}()}} & (11.7\%) & (exact match)\\
                                 & \texttt{{\color{Brown}DFSConfigKeys}.{\color{darkblue}DFS}} & (4.5\%) \\
                                 & \texttt{{\color{Brown}zkfcUgi}.{\color{RoyalPurple}getUserName}()} & (2.6\%) & (tree-match) \\
                                 & \texttt{{\color{Brown}zkfcUgi}.{\color{RoyalPurple}getUser}()} & (1.7\%) & (tree-match)\\
                                 & \texttt{{\color{Brown}zkfcUgi}.{\color{RoyalPurple}getUser}Id()} & (0.6\%) & (tree-match)\\
                                 
\bottomrule
\end{tabular}
\caption*{\emph{Entirely copied tokens} are marked in {\color{Brown}brown}; unknown \emph{copied subtokens} are marked in {\color{darkblue}blue}; \emph{in-vocabulary} subtokens are marked in black; subtokens that are \emph{both in-vocabulary and copied} from context are marked in {\color{RoyalPurple}purple}.}
\end{subfigure}

\caption{A Java \aeg{} example from our test set along with the predictions of our model. 
The predictions of the baselines are shown in \Cref{appendix_java_figure_ugi} below.}
\label{fig:ugi-example}
\end{figure*}
 
\section{Example: Usefulness of Copy Mechanism}


As shown in \Cref{sec:ablation}, the ability to copy is crucial for the \fulltask{} task, because of the repetitive use of identifiers and symbols in programs.
\Cref{fig:ugi-example} shows a representative example for the necessity of the copy mechanism: generating the ground truth \scode{zkfcUgi.getShortUserName()} is feasible \emph{only} thanks to the copy mechanism, since \scode{zkfc} is obviously an UNK subtoken which was not observed in the training data.

In this case, since both \scode{zkfcUgi} and \scode{getShortUserName} appear in context, both were copied as \emph{entire tokens}, rather than generated using subtokens. This  example also shows how the ability to copy \emph{entire tokens} ease the generation process by reducing the number of target symbols (our SLM model is able to copy and combine single subtokens as well).

\section{Java Examples}
\label{appendix:java_examples}
\Crefrange{appendix_java_figure_taskid_first}{appendix_java_figure_last}
contain examples from our test set for the \fulltask{} task in Java, along with the prediction of our model and some of the baselines. The highlighted expressions are the true references that should be generated. Indentation and line breaks may have been altered for typesetting reasons.

\section{C\# Examples}
\label{appendix:csharp_examples}
\Crefrange{appendix_csharp_limited_figure_first}{appendix_csharp_limited_figure_last}
contain examples from our test set for the \restrictgen{} task in C\#
along with the prediction of our model some of the baselines. The highlighted expressions are the true references that should be generated. Indentation and line breaks may have been altered for typesetting reasons.

\begin{figure}[t]
\begin{subfigure}[t!]{1.0\textwidth}
\vspace{8mm}
\begin{minted}[fontsize=\footnotesize, frame=single,framesep=2pt,escapeinside=||]{Java}
private C findCounter(T key) {
    int i = key.ordinal();
    if (counters[i] == null) {
        counters[i] = newCounter(key);
    }
    return |\greenbox{(C) counters[i]}|;
}
\end{minted}
\end{subfigure}
\\
\vspace{1mm}

\begin{subfigure}[]{0.9\textwidth}
\footnotesize
\centering
\begin{tabular}{|l|lr|}
\hline
\bf{Model}  & \bf{Prediction} &  \\
\Xhline{4\arrayrulewidth}
True ref:                            & \texttt{(C) counters[i]} & \\
\Xhline{4\arrayrulewidth}
\multirow{3}{*}{SLM (this work)}             & \texttt{\textbf{(C) counters[i]}} & (71.6\%) \\
                                 & \texttt{(C) this} & (6.3\%) \\
                                 & \texttt{counters[i]} & (4.8\%) \\
\Xhline{4\arrayrulewidth}
\multirow{3}{*}{Transformer$_{\text{base}}$ +copy}             & \texttt{(C) this} & \\
                                 & \texttt{\textbf{(C) counters[i]}} & \\
                                 & \texttt{(C) counters} &\\
\hline
\multirow{3}{*}{BiLSTM$\rightarrow$LSTM +copy}     & \texttt{(C) this} & \\
                                 & \texttt{\textbf{(C) counters[i]}}  & \\
                                 & \texttt{counters[i]} & \\
\hline
\multirow{3}{*}{Seq2tree +copy}             & \texttt{\textbf{(C) counters[i]}} & \\
                                 & \texttt{(C) counters[i].ordinal()}  & \\
                                 & \texttt{(C) counters.get(i)} & \\
\hline
\end{tabular}
\vspace{2mm}
\end{subfigure} \\\vspace{2mm}
\noindent\rule{\textwidth}{1pt}

\begin{subfigure}[t!]{1.0\textwidth}
\vspace{2mm}
\begin{minted}[fontsize=\footnotesize, frame=single,framesep=2pt,escapeinside=||]{Java}
private void handleTaskFinishedEvent(TaskFinishedEvent event) {
  TaskInfo taskInfo = info.tasksMap.get(|\greenbox{event.getTaskId()}|);
  taskInfo.counters = event.getCounters();
  taskInfo.finishTime = event.getFinishTime();
  taskInfo.status = TaskStatus.State.SUCCEEDED.toString();
  taskInfo.successfulAttemptId = event.getSuccessfulTaskAttemptId();
}
\end{minted}
\end{subfigure}
\\
\vspace{1mm}

\begin{subfigure}[]{0.9\textwidth}
\footnotesize
\centering
\begin{tabular}{|l|lr|}
\hline
\bf{Model}  & \bf{Prediction} &  \\
\Xhline{4\arrayrulewidth}
True ref:                            & \texttt{event.getTaskId()} & \\
\Xhline{4\arrayrulewidth}
\multirow{5}{*}{SLM (this work)}             & \texttt{event.getTaskName()} & (8.8\%) \\
                                 & \texttt{event.getId()} & (8.2\%) \\
                                 & \texttt{event.getTask()} & (3.4\%) \\
                                 & \texttt{event.getName()} & (3.3\%) \\
                                 & \texttt{\textbf{event.getTaskId()}} & (3.3\%) \\
\Xhline{4\arrayrulewidth}
\multirow{3}{*}{Transformer$_{\text{base}}$ +copy}             & \texttt{event.getTaskInfo()} & \\
                                 & \textbf{\texttt{event.getTaskId()}} & \\
                                 & \texttt{event.getId()} &\\
                                 & \texttt{event.getTask()} &\\
                                 & \texttt{taskInfo.getTaskId()()} &\\
\hline
\multirow{3}{*}{BiLSTM$\rightarrow$LSTM +copy}     & \texttt{event.name} & \\
                                 & \texttt{event.type}  & \\
                                 & \texttt{event.getId()} & \\
                                 & \texttt{event.id} & \\
                                 & \texttt{event.getKey()} & \\
\hline
\multirow{3}{*}{Seq2tree +copy}             & \texttt{event.getId()} & \\
                                 & \texttt{event.getPath()}  & \\
                                 & \texttt{event.getDescription()} & \\
                                 & \texttt{event.getTaskName()} & \\
                                 & \texttt{event.getTaskName(}  &(\emph{Syntax error})  \\
\hline
\end{tabular}
\end{subfigure} \\\vspace{2mm}

\caption{Java examples from our test set along with the predictions of our model and the baselines.}
\label{appendix_java_figure_taskid_first}
\end{figure} 
\begin{figure}[t]
\begin{subfigure}[t!]{1.0\textwidth}
\vspace{8mm}
\begin{minted}[fontsize=\footnotesize, frame=single,framesep=2pt,escapeinside=||]{Java}
private static void log(String value) {                                                                                                    
    if (value!= null && |\greenbox{value.length() > 55}|)                                                                                                 
        value = value.substring(0, 55) + "...";                                                                                                  
    LOG.info(value);                                                                                                                         
}
\end{minted}
\end{subfigure}
\\
\vspace{1mm}

\begin{subfigure}[]{0.9\textwidth}
\footnotesize
\centering
\begin{tabular}{|l|lr|}
\hline
\bf{Model}  & \bf{Prediction} &  \\
\Xhline{4\arrayrulewidth}
True ref:                            & \texttt{value.length() > 55} & \\
\Xhline{4\arrayrulewidth}
\multirow{3}{*}{SLM (this work)}             & \texttt{value.length() > 0} & (9.6\%) \\
                                 & \texttt{\textbf{value.length() > 55}} & (7.3\%) \\
                                 & \texttt{value.startsWith("...")} & (1.8\%) \\
\Xhline{4\arrayrulewidth}
\multirow{3}{*}{Transformer$_{\text{base}}$ +copy}             & \texttt{\textbf{value.length() > 55}} & \\
                                 & \texttt{value.length() > 0)} & \\
                                 & \texttt{value.length() > 1} &\\
\hline
\multirow{3}{*}{BiLSTM$\rightarrow$LSTM +copy}     & \texttt{\textbf{value.length() > 55}} & \\
                                 & \texttt{value.startsWith("")}  & \\
                                 & \texttt{value.startsWith("...")} & \\
\hline
\multirow{3}{*}{Seq2tree +copy}             & \texttt{value.length() 55} &(\emph{Syntax error}) \\
                                 & \texttt{value.endsWith("info")}  & \\
                                 & \texttt{value.length() 55} &(\emph{Syntax error})\\
\hline
\end{tabular}
\vspace{2mm}
\end{subfigure} \\\vspace{2mm}
\noindent\rule{\textwidth}{1pt}

\begin{subfigure}[t!]{1.0\textwidth}
\vspace{2mm}
\begin{minted}[fontsize=\footnotesize, frame=single,framesep=2pt,escapeinside=||]{Java}
private List<INode> initChildren() {                                                                                                   
  if (children == null) {                                                                                                              
    final ChildrenDiff combined = new ChildrenDiff();                                                                                  
    for (DirectoryDiff d = DirectoryDiff.this; d != null; |\greenbox{d = d.getPosterior()}|) {                                                                                                        
      combined.combinePosterior(d.diff, null);                                                                                         
    }                                                                                                                                  
    children = combined.apply2Current(ReadOnlyList.Util.asList(                                                                        
        currentDir.getChildrenList(Snapshot.CURRENT_STATE_ID)));                                                                       
  }                                                                                                                                    
  return children;                                                                                                                     
}   
\end{minted}
\end{subfigure}
\\
\vspace{1mm}

\begin{subfigure}[]{0.9\textwidth}
\footnotesize
\centering
\begin{tabular}{|l|lr|}
\hline
\bf{Model}  & \bf{Prediction} &  \\
\Xhline{4\arrayrulewidth}
True ref:                            & \texttt{d = d.getPosterior()} & \\
\Xhline{4\arrayrulewidth}
\multirow{5}{*}{SLM (this work)}             & \texttt{d = d.getParent()} & (18.8\%) \\
                                 & \texttt{d = d.getChildrenList()} & (14.9\%) \\
                                 & \texttt{d = d} & (4.5\%) \\
                                 & \texttt{d = combined} & (2.5\%) \\
                                 & \texttt{\textbf{d = d.getPosterior()}} & (1.8\%) \\
\Xhline{4\arrayrulewidth}
\multirow{3}{*}{Transformer$_{\text{base}}$ +copy}             & \texttt{d = d} & \\
                                 & \texttt{d = d.diff} & \\
                                 & \texttt{d = d.getChildren()} &\\
                                 & \texttt{d = d.currentDir} &\\
                                 & \texttt{d = d.currentStateId} &\\
\hline
\multirow{3}{*}{BiLSTM$\rightarrow$LSTM +copy}     & \texttt{--d} & \\
                                 & \texttt{d = d}  & \\
                                 & \texttt{d = d.getParent()} & \\
                                 & \texttt{d = d.next} & \\
                                 & \texttt{d = d.get()} & \\
\hline
\multirow{3}{*}{Seq2tree +copy}             & \texttt{d d.next} &(\emph{Syntax error})  \\
                                  & \texttt{d d.parent} &(\emph{Syntax error})  \\
                                  & \texttt{d d.getParent()} &(\emph{Syntax error})  \\
                                  & \texttt{d d.getChildren()} &(\emph{Syntax error})  \\
                                  & \texttt{d d.getRoot()} &(\emph{Syntax error})  \\
\hline
\end{tabular}
\end{subfigure} \\\vspace{2mm}

\caption{Java examples from our test set along with the predictions of our model and the baselines.}
\label{appendix_java_figure_value_length}
\end{figure} 

\begin{figure}[t]
\begin{subfigure}[t!]{1.0\textwidth}
\vspace{8mm}
\begin{minted}[fontsize=\footnotesize, frame=single,framesep=2pt,escapeinside=||]{Java}
public float getProgress() {
    this.readLock.lock();
    try {
        if (this.currentAttempt != null) {
            return |\greenbox{this.currentAttempt.getProgress()}|;
        }
        return 0;
    } finally {
      this.readLock.unlock();
    }
}
\end{minted}
\end{subfigure}
\\
\vspace{1mm}

\begin{subfigure}[]{0.9\textwidth}
\footnotesize
\centering
\begin{tabular}{|l|lr|}
\hline
\bf{Model}  & \bf{Prediction} &  \\
\Xhline{4\arrayrulewidth}
True ref:                            & \texttt{this.currentAttempt.getProgress()} & \\
\Xhline{4\arrayrulewidth}
\multirow{3}{*}{SLM (this work)}             & \texttt{this.currentAttempt.getCount()} & (31.3\%) \\
                                 & \texttt{-1} & (30.6\%) \\
                                 & \texttt{this.currentAttempt.get()} & (1.5\%) \\
                                 & \texttt{this.currentAttempt.getTime()} & (1.2\%) \\
                                 & \texttt{\textbf{this.currentAttempt.getProgress()}} & (0.9\%) \\
\Xhline{4\arrayrulewidth}
\multirow{3}{*}{Transformer$_{\text{base}}$ +copy}             & \texttt{\textbf{this.currentAttempt.getProgress()}} & \\
                                 & \texttt{this.currentAttempt.floatValue()} & \\
                                 & \texttt{this.currentAttempt.getFloat()} &\\
                                 & \texttt{this.currentAttempt.get()} &\\
                                 & \texttt{this.currentAttempt.getTime()} &\\
\hline
\multirow{3}{*}{BiLSTM$\rightarrow$LSTM +copy}     & \texttt{\textbf{this.currentAttempt.getProgress()}} & \\
                                 & \texttt{this.currentAttempt.float()}  & \\
                                 & \texttt{this.currentAttempt.get()} & \\
                                 & \texttt{this.currentAttempt.size()} & \\
                                 & \texttt{this.currentAttempt.compute()} & \\
\hline
\multirow{3}{*}{Seq2tree +copy}             & \texttt{\textbf{this.currentAttempt.getProgress()}} & \\
                                 & \texttt{this.currentAttempt.floatValue()}  & \\
                                 & \texttt{this.currentAttempt.get()} & \\
                                 & \texttt{this.currentAttempt.getValue()} & \\
                                 & \texttt{(float)this.currentAttempt.size()} & \\\hline 
\end{tabular}
\vspace{2mm}
\end{subfigure} \\\vspace{2mm}
\noindent\rule{\textwidth}{1pt}
\begin{subfigure}[t!]{1.0\textwidth}
\vspace{2mm}
\begin{minted}[fontsize=\footnotesize, frame=single,framesep=2pt,escapeinside=||]{Java}
public int compareTo(LongWritable o) {
    long thisValue = this.value;
    long thatValue = o.value;
    return (thisValue < thatValue ? -1 : (|\greenbox{thisValue == thatValue ? 0 : 1}|));
}
\end{minted}
\end{subfigure}
\\
\vspace{1mm}

\begin{subfigure}[]{0.9\textwidth}
\footnotesize
\centering
\begin{tabular}{|l|lr|}
\hline
\bf{Model}  & \bf{Prediction} &  \\
\Xhline{4\arrayrulewidth}
True ref:                            & \texttt{thisValue == thatValue ? 0 : 1} & \\
\Xhline{4\arrayrulewidth}
\multirow{5}{*}{SLM (this work)}             & \texttt{thisValue == thisValue ? 0 : 1} & (16.3\%) \\
                                 & \texttt{\textbf{thisValue == thatValue ? 0 : 1}} & (11.0\%) \\
                                 & \texttt{thisValue == value ? 0 : 1} & (9.5\%) \\
\Xhline{4\arrayrulewidth}
\multirow{3}{*}{Transformer$_{\text{base}}$ +copy}             & \texttt{thatValue >> thatValue} & \\
                                 & \texttt{thatValue > thatValue ? 1 : 0} & \\
                                 & \texttt{thatValue > thatValue} &\\
\hline
\multirow{3}{*}{BiLSTM$\rightarrow$LSTM +copy}     & \texttt{thisValue - thatValue} & \\
                                 & \texttt{thatValue \& thatValue}  & \\
                                 & \texttt{thatValue ? 1 : 0} & \\
\hline
\multirow{3}{*}{Seq2tree +copy}             & \texttt{thisValue thatValue} &(\emph{Syntax error})  \\
                                 & \texttt{thisValue thatValue 0 1}  &(\emph{Syntax error})  \\
                                 & \texttt{thisValue thatValue 1 0} &(\emph{Syntax error})  \\
\hline
\end{tabular}
\end{subfigure} \\\vspace{2mm}

\caption{Java examples from our test set along with the predictions of our model and the baselines.}
\label{appendix_java_figure_taskid}
\end{figure} 

\centering
\begin{figure}[t]
\centering


\begin{subfigure}[t!]{1.0\textwidth}
\vspace{2mm}
\begin{minted}[fontsize=\footnotesize, frame=single,framesep=2pt,escapeinside=||]{Java}
private static String getNameServiceId(
    Configuration conf, String addressKey) {
  String nameserviceId = conf.get(DFS_NAMESERVICE_ID);
  if (nameserviceId != null) {
    return nameserviceId;
  }
  Collection<String> nsIds = getNameServiceIds(conf);
  if (1 == |\greenbox{nsIds.size()}|) {
    return nsIds.toArray(new String[1])[0];
  }
  String nnId = conf.get(DFS_HA_NAMENODE_ID_KEY);
  return 
    getSuffixIDs(conf, addressKey, null, nnId, LOCAL_ADDRESS_MATCHER)[0];
}
\end{minted}
\end{subfigure}
\\
\vspace{1mm}

\begin{subfigure}[]{0.9\textwidth}
\footnotesize
\centering
\begin{tabular}{|l|lr|}
\hline
\bf{Model}  & \bf{Predictions} &  \\
\Xhline{4\arrayrulewidth}
True ref:                            & \texttt{nsIds.size()} & \\
\Xhline{4\arrayrulewidth}
\multirow{3}{*}{SLM (this work)}             & \texttt{\textbf{nsIds.size()}} & (83.7\%) \\
                                 & \texttt{conf.size()} & (3.0\%) \\
                                 & \texttt{getSuffixIDs(conf).length} & (2.5\%) \\
\Xhline{4\arrayrulewidth}
\multirow{3}{*}{Transformer$_{\text{base}}$ +copy}             & \texttt{-1} & \\
                                 & \texttt{ns.size()} & \\
                                 & \texttt{conf.size()} &\\
\hline
\multirow{3}{*}{BiLSTM$\rightarrow$LSTM +copy}     & \texttt{-1} & \\
                                 & \texttt{Integer.MAX\_VALUE}  & \\
                                 & \texttt{conf.size()} & \\
\hline
\multirow{3}{*}{Seq2tree +copy}             & \texttt{1} & \\
                                 & \texttt{\textbf{nsIds.size()}} & \\
                                 & \texttt{stringPool.blank} & \\
\hline
\end{tabular}
\vspace{2mm}
\end{subfigure} \\
\vspace{1mm}

\noindent\rule{\textwidth}{1pt}
\begin{subfigure}[t!]{1.0\textwidth}
\vspace{2mm}
\begin{minted}[fontsize=\footnotesize, frame=single,framesep=2pt,escapeinside=**]{Java}
protected void checkRpcAdminAccess() throws 
    IOException, AccessControlException {
  UserGroupInformation ugi = UserGroupInformation.getCurrentUser();
  UserGroupInformation zkfcUgi = UserGroupInformation.getLoginUser();
  if (adminAcl.isUserAllowed(ugi) || 
    ugi.getShortUserName().equals(*\greenbox{zkfcUgi.getShortUserName()}*)) {
      LOG.info("Allowed RPC access from " + ugi 
        + " at " + Server.getRemoteAddress());
      return;
    }
  String msg = "Disallowed RPC access from " + ugi 
    + " at " + Server.getRemoteAddress() 
    + ". Not listed in " + DFSConfigKeys.DFS_ADMIN;
  LOG.warn(msg);
  throw new AccessControlException(msg);
}
\end{minted}
\end{subfigure}
\\
\vspace{1mm}

\begin{subfigure}[]{0.9\textwidth}
\footnotesize
\centering
\begin{tabular}{|l|lr|}
\hline
\bf{Model}  & \bf{Predictions} &  \\
\Xhline{4\arrayrulewidth}
True ref:                            & \texttt{zkfcUgi.getShortUserName()} & \\
\Xhline{4\arrayrulewidth}
\multirow{3}{*}{SLM (this work)}             & \textbf{\texttt{zkfcUgi.getShortUserName()}} & (11.7\%) \\
                                 & \texttt{DFSConfigKeys.DFS} & (4.5\%) \\
                                 & \texttt{zkfcUgi.getUserName()} & (2.6\%) \\
\Xhline{4\arrayrulewidth}
\multirow{3}{*}{Transformer$_{\text{base}}$ +copy}             & \texttt{server.getRemoteAddress()} & \\
                                 & \texttt{server.getRemoteUserName()} & \\
                                 & \texttt{server.getShortUserName()} &\\
\hline
\multirow{3}{*}{BiLSTM$\rightarrow$LSTM +copy}     & \texttt{server.getUserName()} & \\
                                 & \texttt{zkfcUgi.getUserName()}  & \\
                                 & \texttt{ugiUgi.getUserName()} & \\
\hline
\multirow{3}{*}{Seq2tree +copy}             & \texttt{dfsConfigKeys.dfsAdmin} & \\
                                 & \texttt{zkfc.getUserName()}  & \\
                                 & \texttt{zkfcUgi.getRemoteAddress()} & \\
\hline
\end{tabular}
\end{subfigure} \\
\vspace{8mm}
\noindent\rule{\textwidth}{1pt}
\caption{Java examples from our test set along with the predictions of our model and the baselines.}
\label{appendix_java_figure_ugi}
\end{figure}


\begin{figure}[t]
\begin{subfigure}[t!]{1.0\textwidth}
\vspace{2mm}
\begin{minted}[fontsize=\footnotesize, frame=single,framesep=2pt,escapeinside=||]{Java}
static String replaceSubstitution(
    String base, Pattern from, String to, boolean repeat) {
  Matcher match = from.matcher(base);
  if (repeat) {
    return |\greenbox{match.replaceAll(to)}|;
  } else {
    return match.replaceFirst(to);
  }
}
\end{minted}
\end{subfigure}
\\
\vspace{1mm}

\begin{subfigure}[]{0.9\textwidth}
\footnotesize
\centering
\begin{tabular}{|l|lr|}
\hline
\bf{Model}  & \bf{Prediction} &  \\
\Xhline{4\arrayrulewidth}
True ref:                            & \texttt{match.replaceAll(to)} & \\
\Xhline{4\arrayrulewidth}
\multirow{3}{*}{SLM (this work)}             & \texttt{match.toString()} & (9.0\%) \\
                                 & \texttt{\textbf{match.replaceAll(to)}} & (8.2\%) \\
                                 & \texttt{match.replaceAll(to, from)} & (6.5\%) \\
\Xhline{4\arrayrulewidth}
\multirow{3}{*}{Transformer$_{\text{base}}$ +copy}             & \texttt{match.replaceFirst(to)} & \\
                                 & \texttt{replace.replaceFirst(to)} & \\
                                 & \texttt{matcher.replaceFirst(to)} &\\
\hline
\multirow{3}{*}{BiLSTM$\rightarrow$LSTM +copy}     & \texttt{match.getFirst()} & \\
                                 & \texttt{match.replaceFirst(to)}  & \\
                                 & \texttt{match.replaceFirst(to, to)} & \\
\hline
\multirow{3}{*}{Seq2tree +copy}             & \texttt{match.replaceFirst(base)} & \\
                                 & \texttt{match.replaceFirst(to)}  & \\
                                 & \texttt{match.replaceFirst(repeat)} & \\
\hline
\end{tabular}
\vspace{2mm}
\end{subfigure} \\\vspace{2mm}
\noindent\rule{\textwidth}{1pt}

\begin{subfigure}[t!]{1.02\textwidth}
\vspace{2mm}
\begin{minted}[fontsize=\footnotesize, frame=single,framesep=2pt,escapeinside=||]{Java}
public void responseReceived(ResponseReceivedEvent event) {
  RequestResult result = event.getRequestResult();
  Date startDate = result.getStartDate();
  Date stopDate = result.getStopDate();
  long elapsed = stopDate.getTime() - startDate.getTime();
  synchronized (this) {
    this.lastE2Elatency = elapsed;
  }
  if (|\greenbox{LOG.isDebugEnabled()}|) {
    int statusCode = result.getStatusCode();
    String etag = result.getEtag();
    HttpURLConnection urlConnection = 
        (HttpURLConnection) event.getConnectionObject();
    int contentLength = urlConnection.getContentLength();
    String requestMethod = urlConnection.getRequestMethod();
    long threadId = Thread.currentThread().getId();
    LOG.debug(String.format(
      "SelfThrottlingIntercept:: ResponseReceived: 
      ... threadId=%d, Status=%d, Elapsed(ms)=%d,  
      ... ETAG=%s, contentLength=%d, requestMethod=%s", 
      threadId, statusCode, elapsed, etag, contentLength, requestMethod));
  }
}
\end{minted}
\end{subfigure}
\\
\vspace{1mm}

\begin{subfigure}[]{0.9\textwidth}
\footnotesize
\centering
\begin{tabular}{|l|lr|}
\hline
\bf{Model}  & \bf{Prediction} &  \\
\Xhline{4\arrayrulewidth}
True ref:                            & \texttt{LOG.isDebugEnabled()} & \\
\Xhline{4\arrayrulewidth}
\multirow{3}{*}{SLM (this work)}             & \texttt{elapsed != null} & (32.1\%) \\
                                 & \texttt{\textbf{LOG.isDebugEnabled()}} & (29.0\%) \\
                                 & \texttt{!LOG.isDebugEnabled()} & (2.4\%) \\
\Xhline{4\arrayrulewidth}
\multirow{3}{*}{Transformer$_{\text{base}}$ +copy}             & \texttt{stopDate != null} & \\
                                 & \texttt{result.hasStatusCode()} & \\
                                 & \texttt{result.hasStatusCode() != elapsed} &\\
\hline
\multirow{3}{*}{BiLSTM$\rightarrow$LSTM +copy}     & \texttt{result != null} & \\
                                 & \texttt{elapsed > 0}  & \\
                                 & \texttt{result.getStatusCode() == workflowConstants.STATUS} & \\
\hline
\multirow{3}{*}{Seq2tree +copy}             & \texttt{event.getConnectionObject() instanceof HttpUrlConnection} & \\
                                 & \texttt{startDate != null}  & \\
                                 & \texttt{\textbf{LOG.isDebugEnabled()}} & \\
\hline
\end{tabular}
\end{subfigure} \\\vspace{2mm}

\caption{Java examples from our test set along with the predictions of our model and the baselines.}
\label{appendix_java_figure}
\end{figure}


\begin{figure}[t]
\begin{subfigure}[t!]{1.0\textwidth}
\vspace{2mm}
\begin{minted}[fontsize=\footnotesize, frame=single,framesep=2pt,escapeinside=**]{Java}
private static boolean isNameResolved(InetAddress address) {
  String hostname = *\greenbox{address.getHostName()}*;
  String ip = address.getHostAddress();
  return !hostname.equals(ip) || NetUtils.isLocalAddress(address);
}
\end{minted}
\end{subfigure}
\\
\vspace{1mm}

\begin{subfigure}[]{0.9\textwidth}
\footnotesize
\centering
\begin{tabular}{|l|lr|}
\hline
\bf{Model}  & \bf{Prediction} &  \\
\Xhline{4\arrayrulewidth}
True ref:                            & \texttt{address.getHostName()} & \\
\Xhline{4\arrayrulewidth}
\multirow{3}{*}{SLM (this work)}             & \texttt{\textbf{address.getHostname()}} & (3.5\%) \\
                                 & \textbf{\texttt{address.getHostName()}} & (2.0\%) \\
                                 & \texttt{inetAddress.getByName(address.getAddress())} & (0.7\%) \\
\Xhline{4\arrayrulewidth}
\multirow{3}{*}{Transformer$_{\text{base}}$ +copy}             & \texttt{address.getHostAddress()} & \\
                                 & \texttt{address.getLastElement().getValue()} & \\
                                 & \texttt{address.getAddress()} &\\
\hline
\multirow{3}{*}{BiLSTM$\rightarrow$LSTM +copy}     & \texttt{address.getHostAddress()} & \\
                                 & \texttt{address.getPort()}  & \\
                                 & \texttt{address.getAddress()} & \\
\hline
\multirow{3}{*}{Seq2tree +copy}             & \texttt{address.getHostAddress()} & \\
                                 & \texttt{address.getPort()}  & \\
                                 & \texttt{address.getAddress()} & \\
\hline
\end{tabular}
\vspace{8mm}
\end{subfigure} \\\vspace{2mm}
\noindent\rule{\textwidth}{1pt}

\begin{subfigure}[t!]{1.0\textwidth}
\vspace{2mm}
\begin{minted}[fontsize=\footnotesize, frame=single,framesep=2pt,escapeinside=||]{Java}
private synchronized void initJournals(List<URI> dirs) {
  int minimumRedundantJournals = conf.getInt(
      DFSConfigKeys.DFS_NAMENODE_EDITS_DIR_MINIMUM_KEY,
      DFSConfigKeys.DFS_NAMENODE_EDITS_DIR_MINIMUM_DEFAULT);
  journalSet = new JournalSet(minimumRedundantJournals);
  for (URI u : dirs) {
    boolean required = 
        FSNamesystem.getRequiredNamespaceEditsDirs(conf).contains(u);
    if (|\greenbox{u.getScheme()}|.equals(NNStorage.LOCAL_URI_SCHEME)) {
      StorageDirectory sd = storage.getStorageDirectory(u);
      if (sd != null) {
        journalSet.add(
            new FileJournalManager(conf, sd, storage), 
            required, sharedEditsDirs.contains(u));
      }
    } else {
      journalSet.add(createJournal(u), 
          required, sharedEditsDirs.contains(u));
    }
  }
  if (journalSet.isEmpty()) {
    LOG.error("No edits directories configured!");
  }
}
\end{minted}
\end{subfigure}
\\
\vspace{1mm}

\begin{subfigure}[]{0.9\textwidth}
\footnotesize
\centering
\begin{tabular}{|l|lr|}
\hline
\bf{Model}  & \bf{Prediction} &  \\
\Xhline{4\arrayrulewidth}
True ref:                            & \texttt{u.getScheme()} & \\
\Xhline{4\arrayrulewidth}
\multirow{3}{*}{SLM (this work)}             & \texttt{u.getName()}  & (27.4\%) \\
                                 & \texttt{\textbf{u.getScheme()}}& (13.1\%) \\
                                 & \texttt{u.getVersion()} & (8.2\%) \\
\Xhline{4\arrayrulewidth}
\multirow{3}{*}{Transformer$_{\text{base}}$ +copy}             & \texttt{journalSet.LOCAL\_URI\_SCHEME} & \\
                                 & \texttt{u.getName()} & \\
                                 & \texttt{Boolean.true} &\\
\hline
\multirow{3}{*}{BiLSTM$\rightarrow$LSTM +copy}     & \texttt{u.toString()} & \\
                                 & \texttt{Boolean.true}  & \\
                                 & \texttt{u.getURI()} & \\
\hline
\multirow{3}{*}{Seq2tree +copy}             & \texttt{\textbf{u.getScheme()}} & \\
                                 & \texttt{u.getName()}  & \\
                                 & \texttt{storage.getLocalUriScheme()} & \\
\hline
\end{tabular}
\end{subfigure} \\\vspace{2mm}

\caption{Java examples from our test set along with the predictions of our model and the baselines.}
\label{appendix_java_figure}
\end{figure}


\begin{figure}[t]
\begin{subfigure}[t!]{1.0\textwidth}
\vspace{2mm}
\begin{minted}[fontsize=\footnotesize, frame=single,framesep=2pt,escapeinside=**]{Java}
static EnumSet<FileAttribute> parse(String s) {
  if (s == null || s.length() == 0) {
    return EnumSet.allOf(FileAttribute.class);
  }
  EnumSet<FileAttribute> set = EnumSet.noneOf(FileAttribute.class);
  FileAttribute[] attributes = values();
  for (char c : *\greenbox{s.toCharArray()}*) {
    int i = 0;
    for (; i < attributes.length && c != attributes[i].symbol; i++) ;
    if (i < attributes.length) {
      if (!set.contains(attributes[i])) {
        set.add(attributes[i]);
      } else {
        throw new IllegalArgumentException("There are more than one '" 
            + attributes[i].symbol + "' in " + s);
      }
    } else {
      throw new IllegalArgumentException("'" + c + "' in " 
          + s + " is undefined.");
    }
  }
  return set;
}
\end{minted}
\end{subfigure}
\\
\vspace{1mm}

\begin{subfigure}[]{0.9\textwidth}
\footnotesize
\centering
\begin{tabular}{|l|lr|}
\hline
\bf{Model}  & \bf{Prediction} &  \\
\Xhline{4\arrayrulewidth}
True ref:                            & \texttt{s.toCharArray()} & \\
\Xhline{4\arrayrulewidth}
\multirow{3}{*}{SLM (this work)}             & \texttt{\textbf{s.toCharArray()}} & (22.4\%) \\
                                 & \texttt{attributes[0].value} & (18.5\%) \\
                                 & \texttt{attributes[undefined].length} & (4.6\%) \\
\Xhline{4\arrayrulewidth}
\multirow{3}{*}{Transformer$_{\text{base}}$ +copy}             & \texttt{s.split(" "} & \\
                                 & \texttt{set.split(" ")} & \\
                                 & \texttt{attributes.keySet()} &\\
\hline
\multirow{3}{*}{BiLSTM$\rightarrow$LSTM +copy}     & \texttt{attributes.length} & \\
                                 & \texttt{attributes[0]}  & \\
                                 & \texttt{attributes[0].next} & \\
\hline
\multirow{3}{*}{Seq2tree +copy}             & \texttt{set.toArray()} & \\
                                 & \texttt{\textbf{s.toCharArray()}}  & \\
                                 & \texttt{set.toCharArray()} & \\
\hline
\end{tabular}
\vspace{2mm}
\end{subfigure} \\\vspace{2mm}
\noindent\rule{\textwidth}{1pt}

\begin{subfigure}[t!]{1.0\textwidth}
\vspace{2mm}
\begin{minted}[fontsize=\footnotesize, frame=single,framesep=2pt,escapeinside=||]{Java}
public static Path[] stat2Paths(FileStatus[] stats) {
  if (stats == null)
    return null;
  Path[] ret = new Path[stats.length];
  for (int i = 0; i < stats.length; ++i) {
    ret[i] = |\greenbox{stats[i].getPath()}|;
  }
  return ret;
}
\end{minted}
\end{subfigure}
\\
\vspace{1mm}

\begin{subfigure}[]{0.9\textwidth}
\footnotesize
\centering
\begin{tabular}{|l|lr|}
\hline
\bf{Model}  & \bf{Prediction} &  \\
\Xhline{4\arrayrulewidth}
True ref:                            & \texttt{stats[i].getPath()} & \\
\Xhline{4\arrayrulewidth}
\multirow{3}{*}{SLM (this work)}             & \texttt{\textbf{stats[i].getPath()}} & (25.2\%) \\
                                 & \texttt{Path(stats[i])} & (3.3\%) \\
                                 & \texttt{new Path(stats[i], charset)} & (2.5\%) \\
                            
\Xhline{4\arrayrulewidth}
\multirow{3}{*}{Transformer$_{\text{base}}$ +copy}             & \texttt{stats[i]} & \\
                                 & \texttt{\textbf{stats[i].getPath()}} & \\
                                 & \texttt{new Path(stats[i])} &\\
\hline
\multirow{3}{*}{BiLSTM$\rightarrow$LSTM +copy}     & \texttt{stats[i]} & \\
                                 & \texttt{new Path(stats[i])}  & \\
                                 & \texttt{stats[i].toString()} & \\
\hline
\multirow{3}{*}{Seq2tree +copy}             & \texttt{stats[i]} & \\
                                 & \texttt{new Path(stats[i])}  & \\
                                 & \texttt{stat(stats[i])} & \\
\hline
\end{tabular}
\end{subfigure} \\\vspace{2mm}

\caption{Java examples from our test set along with the predictions of our model and the baselines.}
\label{appendix_java_figure}
\end{figure}


\begin{figure}[t]
\begin{subfigure}[t!]{1.0\textwidth}
\vspace{2mm}
\begin{minted}[fontsize=\footnotesize, frame=single,framesep=2pt,escapeinside=||]{Java}
void ensureCurrentDirExists() throws IOException {
  for (
      Iterator<StorageDirectory> it = storage.dirIterator(); 
      it.hasNext(); ) {
    StorageDirectory sd = it.next();
    File curDir = sd.getCurrentDir();
    if (|\greenbox{!curDir.exists()}| && !curDir.mkdirs()) {
      throw new IOException("Could not create directory " + curDir);
    }
  }
}
\end{minted}
\end{subfigure}
\\
\vspace{1mm}

\begin{subfigure}[]{0.9\textwidth}
\footnotesize
\centering
\begin{tabular}{|l|lr|}
\hline
\bf{Model}  & \bf{Prediction} &  \\
\Xhline{4\arrayrulewidth}
True ref:                            & \texttt{!curDir.exists()} & \\
\Xhline{4\arrayrulewidth}
\multirow{3}{*}{SLM (this work)}             & \texttt{\textbf{!curDir.exists()}} & (29.0\%) \\
                                 & \texttt{curDir != null} & (25.8\%) \\
                                 & \texttt{curDir.exists()} & (24.4\%) \\
\Xhline{4\arrayrulewidth}
\multirow{3}{*}{Transformer$_{\text{base}}$ +copy}             & \texttt{curDir != null} & \\
                                 & \texttt{\textbf{!curDir.exists()}} & \\
                                 & \texttt{curDir.exists()} &\\
\hline
\multirow{3}{*}{BiLSTM$\rightarrow$LSTM +copy}     & \texttt{curDir != null} & \\
                                 & \texttt{curDir.exists()}  & \\
                                 & \texttt{sd != null} & \\
\hline
\multirow{3}{*}{Seq2tree +copy}             & \texttt{curDir != null} & \\
                                 & \texttt{curDir.exists()}  & \\
                                 & \texttt{\textbf{!curDir.exists()}} & \\
\hline
\end{tabular}
\vspace{2mm}
\end{subfigure} \\\vspace{2mm}
\noindent\rule{\textwidth}{1pt}

\begin{subfigure}[t!]{1.0\textwidth}
\vspace{2mm}
\begin{minted}[fontsize=\footnotesize, frame=single,framesep=2pt,escapeinside=||]{Java}
public static byte[] getXAttr(final Map<?, ?> json, final String name) 
    throws IOException {
  if (json == null) {
    return null;
  }
  Map<String, byte[]> xAttrs = toXAttrs(json);
  if (xAttrs != null) {
    return |\greenbox{xAttrs.get(name)}|;
  }
  return null;
}
\end{minted}
\end{subfigure}
\\
\vspace{1mm}

\begin{subfigure}[]{0.9\textwidth}
\footnotesize
\centering
\begin{tabular}{|l|lr|}
\hline
\bf{Model}  & \bf{Prediction} &  \\
\Xhline{4\arrayrulewidth}
True ref:                            & \texttt{xAttrs.get(name)} & \\
\Xhline{4\arrayrulewidth}
\multirow{3}{*}{SLM (this work)}             & \texttt{\textbf{xAttrs.get(name)}} & (28.2\%) \\
                                 & \texttt{xAttrs.get(xAttrs)} & (5.8\%) \\
                                 & \texttt{xAttrs.toByteArray()} & (4.4\%) \\
\Xhline{4\arrayrulewidth}
\multirow{3}{*}{Transformer$_{\text{base}}$ +copy}             & \texttt{\texttt{\textbf{xAttrs.get(name)}}} & \\
                                 & \texttt{xAttrs.toByteArray()} & \\
                                 & \texttt{new byte[0]} &\\
\hline
\multirow{3}{*}{BiLSTM$\rightarrow$LSTM +copy}     & \texttt{xAttrs.getBytes()} & \\
                                 & \texttt{new byte[0]}  & \\
                                 & \texttt{xAttrs.toByteArray()} & \\
\hline
\multirow{3}{*}{Seq2tree +copy}             & \texttt{\textbf{xAttrs.get(name)}} & \\
                                 & \texttt{xAttrs.get()}  & \\
                                 & \texttt{xAttrs.get(0)} & \\
\hline
\end{tabular}
\end{subfigure} \\\vspace{2mm}

\caption{Java examples from our test set along with the predictions of our model and the baselines.}
\label{appendix_java_figure}
\end{figure}


\begin{figure}[t]
\begin{subfigure}[t!]{1.0\textwidth}
\vspace{2mm}
\begin{minted}[fontsize=\footnotesize, frame=single,framesep=2pt,escapeinside=**]{Java}
private void setFlag(long flag) {
  long prev;
  do {
    prev = unsafe.getLongVolatile(null, this.slotAddress);
    if (*\greenbox{(prev & flag)}* != 0) {
      return;
    }
  } while (!unsafe.compareAndSwapLong(
              null, this.slotAddress, prev, prev | flag));
}
\end{minted}
\end{subfigure}
\\
\vspace{1mm}

\begin{subfigure}[]{0.9\textwidth}
\footnotesize
\centering
\begin{tabular}{|l|lr|}
\hline
\bf{Model}  & \bf{Prediction} &  \\
\Xhline{4\arrayrulewidth}
True ref:                            & \texttt{(prev \& flag)} & \\
\Xhline{4\arrayrulewidth}
\multirow{3}{*}{SLM (this work)}             & \texttt{\textbf{(prev \& flag)}} & (8.9\%) \\
                                 & \texttt{(prev \& flagSlot)} & (5.4\%) \\
                                 & \texttt{unsafe.get(prev)} & (5.0\%) \\
\Xhline{4\arrayrulewidth}
\multirow{3}{*}{Transformer$_{\text{base}}$ +copy}             & \texttt{\textbf{(prev \& flag)}} & \\
                                 & \texttt{(prev | flag)} & \\
                                 & \texttt{unsafe.compareTo(prev)} &\\
\hline
\multirow{3}{*}{BiLSTM$\rightarrow$LSTM +copy}     & \texttt{prev} & \\
                                 & \texttt{prev + 1}  & \\
                                 & \texttt{prev - 1} & \\
\hline
\multirow{3}{*}{Seq2tree +copy}             & \texttt{unsafe prev flag} &(\emph{Syntax error})  \\
                                 & \texttt{(volatile prev unsafe.get())}  &(\emph{Syntax error}) \\
                                 & \texttt{(volatile prev unsafe.getLongVolatile(null, prev))}  &(\emph{Syntax error}) \\
\hline
\end{tabular}
\vspace{2mm}
\end{subfigure} \\\vspace{2mm}
\noindent\rule{\textwidth}{1pt}

\begin{subfigure}[t!]{0.90\textwidth}
\begin{minted}[fontsize=\footnotesize, frame=single,framesep=2pt,escapeinside=**]{Java}
public synchronized void setInput(byte[] b, int off, int len) {
  if (b == null) {
    throw new NullPointerException();
  }
  if (off < 0 || *\greenbox{len < 0}* || off > b.length - len) {
    throw new ArrayIndexOutOfBoundsException();
  }
  finished = false;
  if (len > uncompressedDirectBuf.remaining()) {
    this.userBuf = b;
    this.userBufOff = off;
    this.userBufLen = len;
  } else {
    ((ByteBuffer) uncompressedDirectBuf).put(b, off, len);
    uncompressedDirectBufLen = uncompressedDirectBuf.position();
  }
  bytesRead += len;
}
\end{minted}
\end{subfigure}
\\
\vspace{1mm}

\begin{subfigure}[]{0.9\textwidth}
\footnotesize
\centering
\begin{tabular}{|l|lr|}
\hline
\bf{Model}  & \bf{Predictions} &  \\
\Xhline{4\arrayrulewidth}
True ref:                            & \texttt{len < 0} & \\
\Xhline{4\arrayrulewidth}
\multirow{3}{*}{SLM (this work)}             & \textbf{\texttt{len < 0}} & (41.3\%) \\
                                 & \texttt{off > b.length} & (23.4\%) \\
                                 & \texttt{len > b.length} & (14.1\%) \\
\Xhline{4\arrayrulewidth}
\multirow{3}{*}{Transformer$_{\text{base}}$ +copy}             & \texttt{off < 0} & \\
                                 & \texttt{\textbf{len < 0}} & \\
                                 & \texttt{b == null} &\\
\hline
\multirow{3}{*}{BiLSTM$\rightarrow$LSTM +copy}     & \texttt{off < 0} & \\
                                 & \texttt{\textbf{len < 0}} & \\
                                 & \texttt{b == null} & \\
\hline
\multirow{3}{*}{Seq2tree +copy}     & \texttt{off < 0} & \\
                                 & \texttt{\textbf{len < 0}} & \\
                                 & \texttt{0 < off} & \\
\hline
\end{tabular}
\end{subfigure} \\
\caption{Java examples from our test set along with the predictions of our model and the baselines.}
\label{appendix_java_figure}
\end{figure}



\begin{figure}[t]
\begin{subfigure}[t!]{1.0\textwidth}
\vspace{2mm}
\begin{minted}[fontsize=\footnotesize, frame=single,framesep=2pt,escapeinside=||]{Java}
private int readData(byte[] buf, int off, int len) throws IOException {
  int bytesRead = 0;
  while (bytesRead < len) {
    int n = IOUtils.wrappedReadForCompressedData(
         in, buf, |\greenbox{off + bytesRead}|, len - bytesRead);
    if (n < 0) {
      return bytesRead;
    }
    bytesRead += n;
  }
  return len;
}
\end{minted}
\end{subfigure}
\\
\vspace{1mm}

\begin{subfigure}[]{0.9\textwidth}
\footnotesize
\centering
\begin{tabular}{|l|lr|}
\hline
\bf{Model}  & \bf{Prediction} &  \\
\Xhline{4\arrayrulewidth}
True ref:                            & \texttt{off + bytesRead} & \\
\Xhline{4\arrayrulewidth}
\multirow{3}{*}{SLM (this work)}             & \texttt{bytesRead - bytesRead} & (35.0\%) \\
                                 & \texttt{\textbf{off + bytesRead}} & (14.1\%) \\
                                 & \texttt{off - bytesRead} & (9.4\%) \\
\Xhline{4\arrayrulewidth}
\multirow{3}{*}{Transformer$_{\text{base}}$ +copy}             & \texttt{off - bytesRead} & \\
                                 & \texttt{off + len} & \\
                                 & \texttt{len - bytesRead} &\\
\hline
\multirow{3}{*}{BiLSTM$\rightarrow$LSTM +copy}     & \texttt{-bytesRead} & \\
                                 & \texttt{bytesRead++}  & \\
                                 & \texttt{bytesRead - bytesRead} & \\
\hline
\multirow{3}{*}{Seq2tree +copy}             & \texttt{compressed bytesRead} & (\emph{Syntax error}) \\
                                 & \texttt{\textbf{off + bytesRead}}  & \\
                                 & \texttt{len - bytesRead} & \\
\hline
\end{tabular}
\vspace{2mm}
\end{subfigure} \\\vspace{2mm}
\noindent\rule{\textwidth}{1pt}

\begin{subfigure}[t!]{1.01\textwidth}
\vspace{2mm}
\begin{minted}[fontsize=\footnotesize, frame=single,framesep=2pt,escapeinside=||]{Java}
private Path getPath(int curId, int limitPerDir, Type type) {
  if (curId <= 0) {
    return basePath;
  }
  String name = "";
  switch(type) {
    case FILE:
      name = FILE_PREFIX + new Integer(curId % limitPerDir).toString();
      break;
    case DIRECTORY:
      name = DIR_PREFIX + new Integer(curId % limitPerDir).toString();
      break;
  }
  Path base = getPath((curId / limitPerDir), limitPerDir, Type.DIRECTORY);
  return  |\greenbox{new Path(base, name)}|;
}
\end{minted}
\end{subfigure}
\\
\vspace{1mm}

\begin{subfigure}[]{0.9\textwidth}
\footnotesize
\centering
\begin{tabular}{|l|lr|}
\hline
\bf{Model}  & \bf{Prediction} &  \\
\Xhline{4\arrayrulewidth}
True ref:                            & \texttt{new Path(base, name)} & \\
\Xhline{4\arrayrulewidth}
\multirow{3}{*}{SLM (this work)}             & \texttt{\textbf{new Path(base, name)}} & (6.0\%) \\
                                 & \texttt{new Path(base, name, limitPerDir)} & (2.9\%) \\
                                 & \texttt{new Path(base, name, type)} & (2.8\%) \\
\Xhline{4\arrayrulewidth}
\multirow{3}{*}{Transformer$_{\text{base}}$ +copy}             & \texttt{new Path(base)} & \\
                                 & \texttt{new Path(name)} & \\
                                 & \texttt{getPath(base)} &\\
\hline
\multirow{3}{*}{BiLSTM$\rightarrow$LSTM +copy}     & \texttt{new Path(base)} & \\
                                 & \texttt{new File(base)}  & \\
                                 & \texttt{new Path(base.getPath())} & \\
\hline
\multirow{3}{*}{Seq2tree +copy}             & \texttt{new Path(base)} & \\
                                 & \texttt{new File(base, name)}  & \\
                                 & \texttt{\textbf{new Path(base, name)}} & \\
\hline
\end{tabular}
\end{subfigure} \\\vspace{2mm}
\caption{Java examples from our test set along with the predictions of our model and the baselines.}
\label{appendix_java_figure_last}
\end{figure} 


\begin{figure}[t]
\begin{subfigure}[t!]{1.0\textwidth}
\vspace{2mm}
\begin{minted}[fontsize=\footnotesize, frame=single,framesep=2pt,escapeinside=||]{csharp}
private static IEnumerable<Token> OfSequence(
    this IEnumerable<Token> tokens, Token nameToken, TypeDescriptor info)
{
  var nameIndex = tokens.IndexOf(t => t.Equals(nameToken));
  if (|\greenbox{nameIndex >= 0}|)
  {
    return info.NextValue.MapValueOrDefault(
        _ => info.MaxItems.MapValueOrDefault(
          n => tokens.Skip(nameIndex + 1).Take(n),
               tokens.Skip(nameIndex + 1).TakeWhile(v => v.IsValue())),
        tokens.Skip(nameIndex + 1).TakeWhile(v => v.IsValue()));
  }
  return new Token[] { };
}
\end{minted}
\end{subfigure}
\\
\vspace{1mm}

\begin{subfigure}[]{0.9\textwidth}
\footnotesize
\centering
\begin{tabular}{|l|lr|}
\hline
\bf{Model}  & \bf{Prediction} &  \\
\Xhline{4\arrayrulewidth}
True ref:                            & \texttt{nameIndex >= 0} & \\
\Xhline{4\arrayrulewidth}
\multirow{3}{*}{SLM (this work)}             & \texttt{\textbf{nameIndex >= 0}} & (22.6\%) \\
                                 & \texttt{nameIndex == -1} & (19.1\%) \\
                                 & \texttt{nameIndex > -1} & (13.9\%) \\
\Xhline{4\arrayrulewidth}
\multirow{3}{*}{BiLSTM$\rightarrow$LSTM +copy}     & \texttt{!nameIndex} & \\
                                 & \texttt{nameIndex == -1}  & \\
                                 & \texttt{nameIndex < 0} & \\
\hline
\multirow{3}{*}{\gnntonag{} \cite{brockschmidt2018generative} }             & \texttt{nameIndex == 0} & \\
                                 & \texttt{nameIndex > 0}  & \\
                                 & \texttt{nameIndex < 0} & \\
\hline
\end{tabular}
\vspace{2mm}
\end{subfigure} \\\vspace{2mm}
\noindent\rule{\textwidth}{1pt}

\begin{subfigure}[t!]{1.0\textwidth}
\vspace{2mm}
\begin{minted}[fontsize=\footnotesize, frame=single,framesep=2pt,escapeinside=||]{csharp}
public static IEnumerable<T[]> Group<T>(
    this IEnumerable<T> source, int groupSize)
{
  if (groupSize < 1)
  {
    throw new ArgumentOutOfRangeException(nameof(groupSize));
  }
  T[] group = new T[groupSize];
  int groupIndex = 0;
  foreach (var item in source)
  {
    group[groupIndex++] = item;
    if (|\greenbox{groupIndex == groupSize}|)
    {
      yield return group;
      group = new T[groupSize];
      groupIndex = 0;
    }
  }
}
\end{minted}
\end{subfigure}
\\
\vspace{1mm}

\begin{subfigure}[]{0.9\textwidth}
\footnotesize
\centering
\begin{tabular}{|l|lr|}
\hline
\bf{Model}  & \bf{Prediction} &  \\
\Xhline{4\arrayrulewidth}
True ref:                            & \texttt{groupIndex == groupSize} & \\
\Xhline{4\arrayrulewidth}
\multirow{3}{*}{SLM (this work)}             & \texttt{groupIndex < 0} & (21.4\%) \\
                                 & \texttt{groupIndex == -1} & (10.3\%) \\
                                 & \texttt{groupIndex < groupIndex} & (5.3\%) \\
\Xhline{4\arrayrulewidth}
\multirow{3}{*}{BiLSTM$\rightarrow$LSTM +copy}     & \texttt{group.IsNullOrEmpty()} & \\
                                 & \texttt{groupGroup[groupIndex++]}  & \\
                                 & \texttt{group.EndsWith(group)} & \\
\hline
\multirow{3}{*}{\gnntonag{} \cite{brockschmidt2018generative} }             & \texttt{groupIndex == 0} & \\
                                 & \texttt{groupIndex == 1}  & \\
                                 & \texttt{\textbf{groupIndex == groupSize}} & \\
\hline
\end{tabular}
\end{subfigure} \\\vspace{2mm}

\caption{C\# examples from our test set of the \restrictgen{} task along with the predictions of our model and the baselines.}
\label{appendix_csharp_limited_figure_first}
\end{figure}


\begin{figure}[t]
\begin{subfigure}[t!]{1.0\textwidth}
\vspace{2mm}
\begin{minted}[fontsize=\footnotesize, frame=single,framesep=2pt,escapeinside=||]{csharp}
internal static void AddLine(StringBuilder builder, 
    string value, int maximumLength)
{
  if (builder == null)
  {
    throw new ArgumentNullException(nameof(builder));
  }
  if (value == null)
  {
    throw new ArgumentNullException(nameof(value));
  }
  if (maximumLength < 1)
  {
    throw new ArgumentOutOfRangeException(nameof(value));
  }

  value =  |\greenbox{value.Trim()}|;

  builder.AppendWhen(builder.Length > 0, Environment.NewLine);
  do
  {
    var wordBuffer = 0;
    var words = value.Split(' ');
    for (var i = 0; i < words.Length; i++)
    {
      if (words[i].Length < (maximumLength - wordBuffer))
      {
        builder.Append(words[i]);
        wordBuffer += words[i].Length;
        if ((maximumLength - wordBuffer) > 1 && i != words.Length - 1)
        {
          builder.Append(" ");
          wordBuffer++;
        }
      }
      else if (words[i].Length >= maximumLength && wordBuffer == 0)
      {
        builder.Append(words[i].Substring(0, maximumLength));
        wordBuffer = maximumLength;
        break;
      }
      else break;
    }
    value = value.Substring(Math.Min(wordBuffer, value.Length));
    builder.AppendWhen(value.Length > 0, Environment.NewLine);
  }
  while (value.Length > maximumLength);
  builder.Append(value);
}
\end{minted}
\end{subfigure}
\\
\vspace{1mm}

\begin{subfigure}[]{0.9\textwidth}
\footnotesize
\centering
\begin{tabular}{|l|lr|}
\hline
\bf{Model}  & \bf{Prediction} &  \\
\Xhline{4\arrayrulewidth}
True ref:                            & \texttt{value.Trim()} & \\
\Xhline{4\arrayrulewidth}
\multirow{3}{*}{SLM (this work)}             & \texttt{\textbf{value.Trim()}} & (16.0\%) \\
                                 & \texttt{value.Substring(0, maximumLength)} & (10.9\%) \\
                                 & \texttt{value.Replace(maximumLength, maximumLength} & (10.7\%) \\
\Xhline{4\arrayrulewidth}
\multirow{3}{*}{BiLSTM$\rightarrow$LSTM +copy}     & \texttt{maximumLength - 1} & \\
                                 & \texttt{\textbf{value.Trim()}}  & \\
                                 & \texttt{valueLength++} & \\
\hline
\multirow{3}{*}{\gnntonag{} }             & \texttt{value + <UNK>} & \\
                                 & \texttt{value + maximumLength}  & \\
                                 & \texttt{value.Substring(0, maximumLength)} & \\
\hline
\end{tabular}
\vspace{2mm}
\end{subfigure} \\\vspace{2mm}
\noindent\rule{\textwidth}{1pt}

\caption{C\# examples from our test set of the \restrictgen{} task along with the predictions of our model and the baselines.}
\label{appendix_csharp_limited_figure}
\end{figure}


\begin{figure}[t]
\begin{subfigure}[t!]{1.0\textwidth}
\vspace{2mm}
\begin{minted}[fontsize=\footnotesize, frame=single,framesep=2pt,escapeinside=||]{csharp}
public static string[] TrimStringArray(this IEnumerable<string> array)
{
    return array.Select(item => |\greenbox{item.Trim()}|).ToArray();
}
\end{minted}
\end{subfigure}
\\
\vspace{1mm}

\begin{subfigure}[]{0.9\textwidth}
\footnotesize
\centering
\begin{tabular}{|l|lr|}
\hline
\bf{Model}  & \bf{Prediction} &  \\
\Xhline{4\arrayrulewidth}
True ref:                            & \texttt{item.Trim()} & \\
\Xhline{4\arrayrulewidth}
\multirow{3}{*}{SLM (this work)}             & \texttt{\textbf{item.Trim()}} & (20.1\%) \\
                                 & \texttt{item.ToUpperInvariant()} & (3.5\%) \\
                                 & \texttt{item.ToUpper()} & (1.6\%) \\
\Xhline{4\arrayrulewidth}
\multirow{3}{*}{BiLSTM$\rightarrow$LSTM +copy}     & \texttt{\textbf{item.Trim()}} & \\
                                 & \texttt{item.ToTrim()}  & \\
                                 & \texttt{item.]} & (\emph{Syntax error})\\
\hline
\multirow{3}{*}{\gnntonag{} \cite{brockschmidt2018generative} }             & \texttt{item + <UNK>} & \\
                                 & \texttt{item + item}  & \\
                                 & \texttt{item + 1} & \\
\hline
\end{tabular}
\vspace{2mm}
\end{subfigure} \\\vspace{2mm}
\noindent\rule{\textwidth}{1pt}

\begin{subfigure}[t!]{1.0\textwidth}
\vspace{2mm}
\begin{minted}[fontsize=\footnotesize, frame=single,framesep=2pt,escapeinside=||]{csharp}
public static string Camelize(this string input)
{
    var word = Pascalize(input);
    return |\greenbox{word.Substring(0, 1)}|.ToLower() + |\greenbox{word.Substring(1)}|;
}
\end{minted}
\end{subfigure}
\\
\vspace{1mm}

\begin{subfigure}[]{0.9\textwidth}
\footnotesize
\centering
\begin{tabular}{|l|l|l|}
\hline
\bf{Model}  & \multicolumn{2}{c}{\bf{Prediction}}   \\
\Xhline{4\arrayrulewidth}
True ref:                            & \texttt{word.Substring(0, 1)} & \texttt{word.Substring(1)} \\
\Xhline{4\arrayrulewidth}
\multirow{3}{*}{SLM (this work)}             & \texttt{\textbf{word.Substring(0, 1)}} & \texttt{\textbf{word.Substring(1)}} \\
                                 & \texttt{word.Trim()} & \texttt{wordData.Substring(1)} \\
                                 & \texttt{word.Substring(1)} & \texttt{word.Substring(0, 1)} \\
\Xhline{4\arrayrulewidth}
\multirow{3}{*}{BiLSTM$\rightarrow$LSTM +copy}     & \texttt{input.Replace("\&", " )} &  \texttt{input.Replace("\&", " <UNK> )}\\
                                 & \texttt{input.Replace(1, '')}  &  \texttt{input + "." + input}\\
                                 & \texttt{input.Replace("\&", "")} & \texttt{input.Substring(0, 1)} \\
\hline
\multirow{3}{*}{\gnntonag{}  }             & \texttt{word.CombineWith(<UNK>)} & \texttt{word.CombineWith(<UNK>)} \\
                                 & \texttt{word.Trim()}  & \texttt{word + <UNK>}\\
                                 & \texttt{word.CombineWith(input)} & \texttt{word.Replace(<UNK>, <UNK>)}\\
\hline
\end{tabular}
\end{subfigure} \\\vspace{2mm}

\caption{C\# examples from our test set of the \restrictgen{} task along with the predictions of our model and the baselines.}
\label{appendix_csharp_limited_figure}
\end{figure}


\begin{figure}[t]
\begin{subfigure}[t!]{1.02\textwidth}
\vspace{2mm}
\begin{minted}[fontsize=\footnotesize, frame=single,framesep=2pt,escapeinside=||]{csharp}
public string Truncate(string value, int length, string truncationString, 
    TruncateFrom truncateFrom = TruncateFrom.Right)
{
  if (value == null)
    return null;

  if (value.Length == 0)
    return value;

  if (truncationString == null)
    truncationString = string.Empty;

  if (truncationString.Length > length)
    return truncateFrom == TruncateFrom.Right ? 
      value.Substring(0, length) : value.Substring(value.Length - length);

  var alphaNumericalCharactersProcessed = 0;

  if (value.ToCharArray().Count(char.IsLetterOrDigit) <= length)
    return value;

  if (truncateFrom == TruncateFrom.Left)
  {
    for (var i = value.Length - 1; i > 0; i--)
    {
      if (char.IsLetterOrDigit(value[i]))
        alphaNumericalCharactersProcessed++;

      if (alphaNumericalCharactersProcessed + truncationString.Length 
          == length)
        return truncationString + value.Substring(i);
    }         
  }

  for (var i = 0; i < value.Length - truncationString.Length; i++)
  {
    if (char.IsLetterOrDigit(value[i]))
      |\greenbox{alphaNumericalCharactersProcessed++}|;

    if (alphaNumericalCharactersProcessed + truncationString.Length 
        == length)
      return value.Substring(0, i + 1) + truncationString;
  }

  return value;
}
\end{minted}
\end{subfigure}
\\
\vspace{1mm}

\begin{subfigure}[]{0.9\textwidth}
\footnotesize
\centering
\begin{tabular}{|l|lr|}
\hline
\bf{Model}  & \bf{Prediction} &  \\
\Xhline{4\arrayrulewidth}
True ref:                            & \texttt{alphaNumericalCharactersProcessed++} & \\
\Xhline{4\arrayrulewidth}
\multirow{3}{*}{SLM (this work)}             & \texttt{\textbf{alphaNumericalCharactersProcessed++}} & (48.1\%) \\
                                 & \texttt{iCount++} & (5.8\%) \\
                                 & \texttt{iIndex++} & (1.6\%) \\
\Xhline{4\arrayrulewidth}
\multirow{3}{*}{BiLSTM$\rightarrow$LSTM +copy}     & \texttt{i++} & \\
                                 & \texttt{truncation++}  & \\
                                 & \texttt{alpha--} & \\
\hline
\multirow{3}{*}{\gnntonag{} }             & \texttt{\textbf{alphaNumericalCharactersProcessed++}} & \\
                                 & \texttt{alphaNumericalCharactersProcessed--}  & \\
                                 & \texttt{--alphaNumericalCharactersProcessed} & \\
\hline
\end{tabular}
\vspace{2mm}
\end{subfigure} \\\vspace{2mm}
\noindent\rule{\textwidth}{1pt}

\caption{C\# examples from our test set of the \restrictgen{} task along with the predictions of our model and the baselines.}
\label{appendix_csharp_limited_figure}
\end{figure}


\begin{figure}[t]
\begin{subfigure}[t!]{1.0\textwidth}
\vspace{2mm}
\begin{minted}[fontsize=\footnotesize, frame=single,framesep=2pt,escapeinside=||]{csharp}
public static int BinarySearch<TItem, TSearch>(
    this IList<TItem> list, TSearch value, 
    Func<TSearch, TItem, int> comparer)
{
  if (list == null)
  {
    throw new ArgumentNullException("list");
  }
  if (comparer == null)
  {
    throw new ArgumentNullException("comparer");
  }

  var lower = 0;
  var upper = list.Count - 1;

  while (lower <= upper)
  {
    var middle = lower + (upper - lower) / 2;
    var comparisonResult = comparer(value, list[middle]);
    if (|\greenbox{comparisonResult < 0}|)
    {
      upper = middle - 1;
    }
    else if (|\greenbox{comparisonResult > 0}|)
    {
      lower = middle + 1;
    }
    else
    {
      return middle;
    }
  }

  return lower;
}
\end{minted}
\end{subfigure}
\\
\vspace{1mm}

\begin{subfigure}[]{0.9\textwidth}
\footnotesize
\centering
\begin{tabular}{|l|l|l|}
\hline
\bf{Model}  & \bf{Prediction} &  \\
\Xhline{4\arrayrulewidth}
True ref:                            & \texttt{comparisonResult < 0} & \texttt{comparisonResult > 0}\\
\Xhline{4\arrayrulewidth}
\multirow{3}{*}{SLM (this work)}             & \texttt{\textbf{comparisonResult < 0}} & \texttt{\textbf{comparisonResult > 0}} \\
                                 & \texttt{comparisonResult > 0} & \texttt{comparisonResult < 0} \\
                                 & \texttt{middle == comparisonResult} & \texttt{comparisonResult == 0} \\
\Xhline{4\arrayrulewidth}
\multirow{3}{*}{BiLSTM$\rightarrow$LSTM +copy}     & \texttt{lowerResult == middle} & \texttt{lower < 0}\\
                                 & \texttt{lowerResult == 0}  & \texttt{lower + "."}\\
                                 & \texttt{lower != middle} & \texttt{lower != middle}\\
\hline
\multirow{3}{*}{\gnntonag{} }             & \texttt{comparisonResult == 0} & \texttt{comparisonResult == 0}\\
                                 & \texttt{comparisonResult > 0}  & \texttt{\textbf{comparisonResult > 0}}\\
                                 & \texttt{\textbf{comparisonResult < 0}}  & \texttt{comparisonResult == middle}\\
\hline
\end{tabular}
\vspace{2mm}
\end{subfigure} \\\vspace{2mm}

\caption{C\# examples from our test set of the \restrictgen{} task along with the predictions of our model and the baselines.}
\label{appendix_csharp_limited_figure}
\end{figure}


\begin{figure}[t]
\begin{subfigure}[t!]{1.0\textwidth}
\vspace{2mm}
\begin{minted}[fontsize=\footnotesize, frame=single,framesep=2pt,escapeinside=**]{csharp}
public override string ToString()
{
  // use reflection to display all the properties that 
  // ... have non default values
  StringBuilder result = new StringBuilder();
  var props = this.GetType().GetTypeInfo().DeclaredProperties;
  result.AppendLine("{");
  foreach (var prop in props)
  {
    if (prop.Name != "Content" && prop.Name != "Subtitle" 
        && prop.Name != "Title" && prop.Name != "UniqueId")
    {
        object value = prop.GetValue(this);
        bool valueIsNull = value == null;
        object defaultValue = Common.GetDefault(prop.PropertyType);
        bool defaultValueIsNull = defaultValue == null;
        if ((valueIsNull != defaultValueIsNull) 
            // one is null when the other isn't
          || (*\greenbox{!valueIsNull}* 
               && (value.ToString() != defaultValue.ToString()))) 
            // both aren't null, so compare as strings
        {
          result.AppendLine(prop.Name + " : " + prop.GetValue(this));
        }
    }
  }
  result.AppendLine("}");
  return result.ToString();
}
\end{minted}
\end{subfigure}
\\
\vspace{1mm}

\begin{subfigure}[]{0.9\textwidth}
\footnotesize
\centering
\begin{tabular}{|l|lr|}
\hline
\bf{Model}  & \bf{Prediction} &  \\
\Xhline{4\arrayrulewidth}
True ref:                            & \texttt{!valueIsNull} & \\
\Xhline{4\arrayrulewidth}
\multirow{3}{*}{SLM (this work)}             & \texttt{\textbf{!valueIsNull}} & (52.4\%) \\
                                 & \texttt{!defaultValueIsNull} & (9.0\%) \\
                                 & \texttt{!valueIsNull.IsNullOrEmpty()} & (3.2\%) \\
\Xhline{4\arrayrulewidth}
\multirow{3}{*}{BiLSTM$\rightarrow$LSTM +copy}     & \texttt{!defaultValueIsNull} & \\
                                 & \texttt{(defaultValueIsNull || value)}  & \\
                                 & \texttt{(defaultValueIsNull || defaultValue)} & \\
\hline
\multirow{3}{*}{\gnntonag{} }             & \texttt{\textbf{!valueIsNull}} & \\
                                 & \texttt{!defaultValueIsNull}  & \\
                                 & \texttt{!!valueIsNull} & \\
\hline
\end{tabular}
\vspace{2mm}
\end{subfigure} \\\vspace{2mm}

\caption{C\# examples from our test set of the \restrictgen{} task along with the predictions of our model and the baselines.}
\label{appendix_csharp_limited_figure}
\end{figure}


\begin{figure}[t]
\begin{subfigure}[t!]{1.0\textwidth}
\vspace{2mm}
\begin{minted}[fontsize=\footnotesize, frame=single,framesep=2pt,escapeinside=||]{csharp}
public TradierOrderResponse PlaceOrder(string accountId,
    TradierOrderClass classification,
    TradierOrderDirection direction,
    string symbol,
    decimal quantity,
    decimal price = 0,
    decimal stop = 0,
    string optionSymbol = "",
    TradierOrderType type = TradierOrderType.Market,
    TradierOrderDuration duration = TradierOrderDuration.GTC)
{
    //Compose the request:
    var request = new RestRequest("accounts/{accountId}/orders");
    request.AddUrlSegment("accountId", accountId.ToString());

    //Add data:
    request.AddParameter("class", GetEnumDescription(classification));
    request.AddParameter("symbol", symbol);
    request.AddParameter("duration", GetEnumDescription(duration));
    request.AddParameter("type", GetEnumDescription(type));
    request.AddParameter("quantity", quantity);
    request.AddParameter("side", GetEnumDescription(direction));

    //Add optionals:
    if (price > 0) request.AddParameter("price", Math.Round(price, 2));
    if (stop > 0) request.AddParameter("stop", Math.Round(stop, 2));
    if (|\greenbox{optionSymbol != {|""|}}|)
        request.AddParameter("option_symbol", optionSymbol);

    //Set Method:
    request.Method = Method.POST;

    return Execute<TradierOrderResponse>(request,
        TradierApiRequestType.Orders);
}
\end{minted}
\end{subfigure}
\\
\vspace{1mm}

\begin{subfigure}[]{0.9\textwidth}
\footnotesize
\centering
\begin{tabular}{|l|lr|}
\hline
\bf{Model}  & \bf{Prediction} &  \\
\Xhline{4\arrayrulewidth}
True ref:                            & \texttt{optionSymbol != ""} & \\
\Xhline{4\arrayrulewidth}
\multirow{3}{*}{SLM (this work)}             & \texttt{\textbf{optionSymbol != ""}} & (5.5\%) \\
                                 & \texttt{optionSymbol == ""} & (4.4\%) \\
                                 & \texttt{optionSymbol.IsNullOrEmpty()} & (1.1\%) \\
\Xhline{4\arrayrulewidth}
\multirow{3}{*}{BiLSTM$\rightarrow$LSTM +copy}     & \texttt{!stopSymbol} & \\
                                 & \texttt{stopSymbol != optionSymbol}  & \\
                                 & \texttt{(stopSymbol " \&\& optionSymbol)} & (\emph{Syntax error})\\
\hline
\multirow{3}{*}{\gnntonag{} }             & \texttt{optionSymbol == <UNK>} & \\
                                 & \texttt{optionSymbol == symbol}  & \\
                                 & \texttt{optionSymbol != symbol} & \\
\hline
\end{tabular}
\vspace{2mm}
\end{subfigure} \\\vspace{2mm}

\caption{C\# examples from our test set of the \restrictgen{} task along with the predictions of our model and the baselines.}
\label{appendix_csharp_limited_figure}
\end{figure}


\begin{figure}[t]
\begin{subfigure}[t!]{1.0\textwidth}
\vspace{2mm}
\begin{minted}[fontsize=\footnotesize, frame=single,framesep=2pt,escapeinside=||]{csharp}
[Test, TestCaseSource("GetLeanDataLineTestParameters")]
public void GetSourceMatchesGenerateZipFilePath(
    LeanDataLineTestParameters parameters)
{
    var source = parameters.Data.GetSource(
        parameters.Config, parameters.Data.Time.Date, false);
    var normalizedSourcePath = new FileInfo(source.Source).FullName;
    var zipFilePath = LeanData.GenerateZipFilePath(
        Globals.DataFolder, parameters.Data.Symbol, 
        parameters.Data.Time.Date, 
        parameters.Resolution, parameters.TickType);
    var normalizeZipFilePath = new FileInfo(zipFilePath).FullName;
    var indexOfHash = normalizedSourcePath.LastIndexOf(
        "#", StringComparison.Ordinal);
    if (indexOfHash > 0)
    {
        normalizedSourcePath = 
            |\greenbox{normalizedSourcePath.Substring(0, indexOfHash)}|;
    }
    Assert.AreEqual(normalizeZipFilePath, normalizedSourcePath);
}
\end{minted}
\end{subfigure}
\\
\vspace{1mm}

\begin{subfigure}[]{0.9\textwidth}
\footnotesize
\centering
\begin{tabular}{|l|lr|}
\hline
\bf{Model}  & \bf{Prediction} &  \\
\Xhline{4\arrayrulewidth}
True ref:                            & \texttt{normalizedSourcePath.Substring(0, indexOfHash)} & \\
\Xhline{4\arrayrulewidth}
\multirow{3}{*}{SLM (this work)}             & \texttt{\textbf{normalizedSourcePath.Substring(0, indexOfHash)}} & (28.3\%) \\
                                 & \texttt{normalizedSourcePath.Substring(1)} & (8.8\%) \\
                                 & \texttt{normalizedSourcePath.Remove(indexOfHash)} & (8.2\%) \\
\Xhline{4\arrayrulewidth}
\multirow{3}{*}{BiLSTM$\rightarrow$LSTM +copy}     & \texttt{indexOfHash + "<UNK>"} & \\
                                 & \texttt{indexOfHash > normalizedOfHash}  & \\
                                 & \texttt{indexOfHash > 0} & \\
\hline
\multirow{3}{*}{\gnntonag{}  }             & \texttt{normalizedSourcePath + normalizeZipFilePath} & \\
                                 & \texttt{normalizedSourcePath + normalizedSourcePath}  & \\
                                 & \texttt{normalizedSourcePath + normalizeZipFilePath + <UNK>} & \\
\hline
\end{tabular}
\vspace{2mm}
\end{subfigure} \\\vspace{2mm}
\noindent\rule{\textwidth}{1pt}

\caption{C\# examples from our test set of the \restrictgen{} task along with the predictions of our model and the baselines.}
\label{appendix_csharp_limited_figure_last}
\end{figure}

\clearpage
\bibliography{bib}

\begin{thebibliography}{46}
\providecommand{\natexlab}[1]{#1}
\providecommand{\url}[1]{\texttt{#1}}
\expandafter\ifx\csname urlstyle\endcsname\relax
  \providecommand{\doi}[1]{doi: #1}\else
  \providecommand{\doi}{doi: \begingroup \urlstyle{rm}\Url}\fi

\bibitem[Aharoni \& Goldberg(2017)Aharoni and Goldberg]{aharoni2017towards}
Aharoni, R. and Goldberg, Y.
\newblock Towards string-to-tree neural machine translation.
\newblock In \emph{Proceedings of the 55th Annual Meeting of the Association
  for Computational Linguistics (Volume 2: Short Papers)}, pp.\  132--140,
  2017.

\bibitem[Allamanis(2019)]{allamanis2018adverse}
Allamanis, M.
\newblock The adverse effects of code duplication in machine learning models of
  code.
\newblock In \emph{Proceedings of the 2019 ACM SIGPLAN International Symposium
  on New Ideas, New Paradigms, and Reflections on Programming and Software},
  pp.\  143--153. ACM, 2019.

\bibitem[Allamanis et~al.(2015)Allamanis, Tarlow, Gordon, and Wei]{bimodal15}
Allamanis, M., Tarlow, D., Gordon, A., and Wei, Y.
\newblock Bimodal modelling of source code and natural language.
\newblock In \emph{International conference on machine learning}, pp.\
  2123--2132, 2015.

\bibitem[Allamanis et~al.(2016)Allamanis, Peng, and Sutton]{conv16}
Allamanis, M., Peng, H., and Sutton, C.
\newblock A convolutional attention network for extreme summarization of source
  code.
\newblock In \emph{International conference on machine learning}, pp.\
  2091--2100, 2016.

\bibitem[Allamanis et~al.(2018)Allamanis, Brockschmidt, and
  Khademi]{allamanis2018learning}
Allamanis, M., Brockschmidt, M., and Khademi, M.
\newblock Learning to represent programs with graphs.
\newblock In \emph{International Conference on Learning Representations}, 2018.

\bibitem[Alon \& Yahav(2020)Alon and Yahav]{alon2020bottleneck}
Alon, U. and Yahav, E.
\newblock On the bottleneck of graph neural networks and its practical
  implications.
\newblock \emph{arXiv preprint arXiv:2006.05205}, 2020.

\bibitem[Alon et~al.(2018)Alon, Zilberstein, Levy, and Yahav]{pigeon}
Alon, U., Zilberstein, M., Levy, O., and Yahav, E.
\newblock A general path-based representation for predicting program
  properties.
\newblock In \emph{Proceedings of the 39th ACM SIGPLAN Conference on
  Programming Language Design and Implementation}, pp.\  404--419, 2018.

\bibitem[Alon et~al.(2019{\natexlab{a}})Alon, Brody, Levy, and
  Yahav]{alon2019code2seq}
Alon, U., Brody, S., Levy, O., and Yahav, E.
\newblock code2seq: Generating sequences from structured representations of
  code.
\newblock In \emph{International Conference on Learning Representations},
  2019{\natexlab{a}}.

\bibitem[Alon et~al.(2019{\natexlab{b}})Alon, Zilberstein, Levy, and
  Yahav]{alon2019code2vec}
Alon, U., Zilberstein, M., Levy, O., and Yahav, E.
\newblock code2vec: Learning distributed representations of code.
\newblock \emph{Proceedings of the ACM on Programming Languages}, 3\penalty0
  (POPL):\penalty0 1--29, 2019{\natexlab{b}}.

\bibitem[Amodio et~al.(2017)Amodio, Chaudhuri, and Reps]{amodio2017neural}
Amodio, M., Chaudhuri, S., and Reps, T.
\newblock Neural attribute machines for program generation.
\newblock \emph{arXiv preprint arXiv:1705.09231}, 2017.

\bibitem[Balog et~al.(2017)Balog, Gaunt, Brockschmidt, Nowozin, and
  Tarlow]{balog2016deepcoder}
Balog, M., Gaunt, A.~L., Brockschmidt, M., Nowozin, S., and Tarlow, D.
\newblock Deepcoder: Learning to write programs.
\newblock In \emph{International Conference on Learning Representations}, 2017.

\bibitem[Bielik et~al.(2016)Bielik, Raychev, and Vechev]{phog16}
Bielik, P., Raychev, V., and Vechev, M.
\newblock Phog: probabilistic model for code.
\newblock In \emph{International Conference on Machine Learning}, pp.\
  2933--2942, 2016.

\bibitem[Brockschmidt et~al.(2019)Brockschmidt, Allamanis, Gaunt, and
  Polozov]{brockschmidt2018generative}
Brockschmidt, M., Allamanis, M., Gaunt, A.~L., and Polozov, O.
\newblock Generative code modeling with graphs.
\newblock In \emph{International Conference on Learning Representations}, 2019.

\bibitem[Brody et~al.(2020)Brody, Alon, and Yahav]{brody2020neural}
Brody, S., Alon, U., and Yahav, E.
\newblock Neural edit completion.
\newblock \emph{arXiv preprint arXiv:2005.13209}, 2020.

\bibitem[Chen et~al.(2018)Chen, Liu, and Song]{chen2018tree}
Chen, X., Liu, C., and Song, D.
\newblock Tree-to-tree neural networks for program translation.
\newblock In \emph{Advances in Neural Information Processing Systems}, pp.\
  2547--2557, 2018.

\bibitem[Devlin et~al.(2017)Devlin, Uesato, Bhupatiraju, Singh, Mohamed, and
  Kohli]{devlin2017robustfill}
Devlin, J., Uesato, J., Bhupatiraju, S., Singh, R., Mohamed, A.-r., and Kohli,
  P.
\newblock Robustfill: Neural program learning under noisy i/o.
\newblock In \emph{International Conference on Machine Learning}, pp.\
  990--998, 2017.

\bibitem[Dong \& Lapata(2018)Dong and Lapata]{dong2018coarse}
Dong, L. and Lapata, M.
\newblock Coarse-to-fine decoding for neural semantic parsing.
\newblock In \emph{Proceedings of the 56th Annual Meeting of the Association
  for Computational Linguistics (Volume 1: Long Papers)}, pp.\  731--742, 2018.

\bibitem[Ellis et~al.(2019)Ellis, Nye, Pu, Sosa, Tenenbaum, and
  Solar-Lezama]{ellis2019write}
Ellis, K., Nye, M., Pu, Y., Sosa, F., Tenenbaum, J., and Solar-Lezama, A.
\newblock Write, execute, assess: Program synthesis with a repl.
\newblock In \emph{Advances in Neural Information Processing Systems}, pp.\
  9169--9178, 2019.

\bibitem[Fernandes et~al.(2019)Fernandes, Allamanis, and
  Brockschmidt]{fernandes2018structured}
Fernandes, P., Allamanis, M., and Brockschmidt, M.
\newblock Structured neural summarization.
\newblock In \emph{International Conference on Learning Representations}, 2019.

\bibitem[Gaunt et~al.(2017)Gaunt, Brockschmidt, Kushman, and
  Tarlow]{gaunt2017differentiable}
Gaunt, A.~L., Brockschmidt, M., Kushman, N., and Tarlow, D.
\newblock Differentiable programs with neural libraries.
\newblock In \emph{International Conference on Machine Learning}, pp.\
  1213--1222, 2017.

\bibitem[Green(1981)]{green1981application}
Green, C.
\newblock Application of theorem proving to problem solving.
\newblock In \emph{Readings in Artificial Intelligence}, pp.\  202--222.
  Elsevier, 1981.

\bibitem[Gu et~al.(2016)Gu, Lu, Li, and Li]{gu2016incorporating}
Gu, J., Lu, Z., Li, H., and Li, V.~O.
\newblock Incorporating copying mechanism in sequence-to-sequence learning.
\newblock In \emph{Proceedings of the 54th Annual Meeting of the Association
  for Computational Linguistics (Volume 1: Long Papers)}, pp.\  1631--1640,
  2016.

\bibitem[Gulwani(2011)]{gulwani2011automating}
Gulwani, S.
\newblock Automating string processing in spreadsheets using input-output
  examples.
\newblock In \emph{Proceedings of the 38th annual ACM SIGPLAN-SIGACT symposium
  on Principles of programming languages}, pp.\  317--330, 2011.

\bibitem[Gulwani et~al.(2017)Gulwani, Polozov, Singh,
  et~al.]{gulwani2017program}
Gulwani, S., Polozov, O., Singh, R., et~al.
\newblock Program synthesis.
\newblock \emph{Foundations and Trends{\textregistered} in Programming
  Languages}, 4\penalty0 (1-2):\penalty0 1--119, 2017.

\bibitem[Iyer et~al.(2018)Iyer, Konstas, Cheung, and
  Zettlemoyer]{iyer2018mapping}
Iyer, S., Konstas, I., Cheung, A., and Zettlemoyer, L.
\newblock Mapping language to code in programmatic context.
\newblock In \emph{Proceedings of the 2018 Conference on Empirical Methods in
  Natural Language Processing}, pp.\  1643--1652, 2018.

\bibitem[Iyer et~al.(2019)Iyer, Cheung, and Zettlemoyer]{iyer2019learning}
Iyer, S., Cheung, A., and Zettlemoyer, L.
\newblock Learning programmatic idioms for scalable semantic parsing.
\newblock In \emph{Proceedings of the 2019 Conference on Empirical Methods in
  Natural Language Processing and the 9th International Joint Conference on
  Natural Language Processing (EMNLP-IJCNLP)}, pp.\  5429--5438, 2019.

\bibitem[Kingma \& Ba(2015)Kingma and Ba]{kingma2014adam}
Kingma, D. and Ba, J.
\newblock Adam: A method for stochastic optimization.
\newblock In \emph{International Conference on Learning Representations}, 2015.

\bibitem[Klein et~al.(2017)Klein, Kim, Deng, Senellart, and Rush]{2017opennmt}
Klein, G., Kim, Y., Deng, Y., Senellart, J., and Rush, A.~M.
\newblock Opennmt: Open-source toolkit for neural machine translation.
\newblock In \emph{Proceedings of ACL 2017, System Demonstrations}, pp.\
  67--72, 2017.

\bibitem[Ling et~al.(2016)Ling, Blunsom, Grefenstette, Hermann,
  Ko{\v{c}}isk{\`y}, Wang, and Senior]{ling2016latent}
Ling, W., Blunsom, P., Grefenstette, E., Hermann, K.~M., Ko{\v{c}}isk{\`y}, T.,
  Wang, F., and Senior, A.
\newblock Latent predictor networks for code generation.
\newblock In \emph{Proceedings of the 54th Annual Meeting of the Association
  for Computational Linguistics (Volume 1: Long Papers)}, pp.\  599--609, 2016.

\bibitem[Luong et~al.(2015)Luong, Pham, and Manning]{luong15}
Luong, M.-T., Pham, H., and Manning, C.~D.
\newblock Effective approaches to attention-based neural machine translation.
\newblock In \emph{Proceedings of the 2015 Conference on Empirical Methods in
  Natural Language Processing}, pp.\  1412--1421, 2015.

\bibitem[Maddison \& Tarlow(2014)Maddison and Tarlow]{maddison2014}
Maddison, C. and Tarlow, D.
\newblock Structured generative models of natural source code.
\newblock In \emph{International Conference on Machine Learning}, pp.\
  649--657, 2014.

\bibitem[Murali et~al.(2018)Murali, Qi, Chaudhuri, and
  Jermaine]{murali2018bayou}
Murali, V., Qi, L., Chaudhuri, S., and Jermaine, C.
\newblock Neural sketch learning for conditional program generation.
\newblock In \emph{International Conference on Learning Representations}, 2018.

\bibitem[Parisotto et~al.(2017)Parisotto, Mohamed, Singh, Li, Zhou, and
  Kohli]{parisotto2016neuro}
Parisotto, E., Mohamed, A.-r., Singh, R., Li, L., Zhou, D., and Kohli, P.
\newblock Neuro-symbolic program synthesis.
\newblock In \emph{International Conference on Learning Representations}, 2017.

\bibitem[Pnueli \& Rosner(1989)Pnueli and Rosner]{pnueli1989synthesis}
Pnueli, A. and Rosner, R.
\newblock On the synthesis of a reactive module.
\newblock In \emph{Proceedings of the 16th ACM SIGPLAN-SIGACT symposium on
  Principles of programming languages}, pp.\  179--190. ACM, 1989.

\bibitem[Polozov \& Gulwani(2015)Polozov and Gulwani]{polozov2015flashmeta}
Polozov, O. and Gulwani, S.
\newblock Flashmeta: a framework for inductive program synthesis.
\newblock In \emph{Proceedings of the 2015 ACM SIGPLAN International Conference
  on Object-Oriented Programming, Systems, Languages, and Applications}, pp.\
  107--126, 2015.

\bibitem[Rabinovich et~al.(2017)Rabinovich, Stern, and Klein]{rabinovich2017}
Rabinovich, M., Stern, M., and Klein, D.
\newblock Abstract syntax networks for code generation and semantic parsing.
\newblock In \emph{Proceedings of the 55th Annual Meeting of the Association
  for Computational Linguistics (Volume 1: Long Papers)}, pp.\  1139--1149,
  2017.

\bibitem[Raychev et~al.(2016)Raychev, Bielik, Vechev, and
  Krause]{raychev2016noisy}
Raychev, V., Bielik, P., Vechev, M., and Krause, A.
\newblock Learning programs from noisy data.
\newblock In \emph{Proceedings of the 43rd Annual ACM SIGPLAN-SIGACT Symposium
  on Principles of Programming Languages}, pp.\  761--774, 2016.

\bibitem[Si et~al.(2019)Si, Yang, Dai, Naik, and Song]{si2018learning}
Si, X., Yang, Y., Dai, H., Naik, M., and Song, L.
\newblock Learning a meta-solver for syntax-guided program synthesis.
\newblock In \emph{International Conference on Learning Representations}, 2019.

\bibitem[Vaswani et~al.(2017)Vaswani, Shazeer, Parmar, Uszkoreit, Jones, Gomez,
  Kaiser, and Polosukhin]{vaswani2017attention}
Vaswani, A., Shazeer, N., Parmar, N., Uszkoreit, J., Jones, L., Gomez, A.~N.,
  Kaiser, {\L}., and Polosukhin, I.
\newblock Attention is all you need.
\newblock In \emph{Advances in Neural Information Processing Systems}, pp.\
  6000--6010, 2017.

\bibitem[Waldinger \& Lee(1969)Waldinger and Lee]{waldinger1969prow}
Waldinger, R.~J. and Lee, R.~C.
\newblock Prow: A step toward automatic program writing.
\newblock In \emph{Proceedings of the 1st international joint conference on
  Artificial intelligence}, pp.\  241--252, 1969.

\bibitem[Xiao et~al.(2016)Xiao, Dymetman, and Gardent]{xiao2016sequence}
Xiao, C., Dymetman, M., and Gardent, C.
\newblock Sequence-based structured prediction for semantic parsing.
\newblock In \emph{Proceedings of the 54th Annual Meeting of the Association
  for Computational Linguistics (Volume 1: Long Papers)}, pp.\  1341--1350,
  2016.

\bibitem[Yin \& Neubig(2017)Yin and Neubig]{yin2017}
Yin, P. and Neubig, G.
\newblock A syntactic neural model for general-purpose code generation.
\newblock In \emph{Proceedings of the 55th Annual Meeting of the Association
  for Computational Linguistics (Volume 1: Long Papers)}, pp.\  440--450, 2017.

\bibitem[Yin et~al.(2019)Yin, Neubig, Allamanis, Brockschmidt, and
  Gaunt]{yin2018learning}
Yin, P., Neubig, G., Allamanis, M., Brockschmidt, M., and Gaunt, A.~L.
\newblock Learning to represent edits.
\newblock In \emph{International Conference on Learning Representations}, 2019.

\bibitem[Young et~al.(2019)Young, Bastani, and Naik]{young2019learning}
Young, H., Bastani, O., and Naik, M.
\newblock Learning neurosymbolic generative models via program synthesis.
\newblock In \emph{International Conference on Machine Learning}, pp.\
  7144--7153, 2019.

\bibitem[Yu et~al.(2018)Yu, Li, Zhang, Zhang, and Radev]{yu2018typesql}
Yu, T., Li, Z., Zhang, Z., Zhang, R., and Radev, D.
\newblock Typesql: Knowledge-based type-aware neural text-to-sql generation.
\newblock In \emph{Proceedings of the 2018 Conference of the North American
  Chapter of the Association for Computational Linguistics: Human Language
  Technologies, Volume 2 (Short Papers)}, pp.\  588--594, 2018.

\bibitem[Zhao et~al.(2019)Zhao, Bieber, Swersky, and Tarlow]{zhao2019neural}
Zhao, R., Bieber, D., Swersky, K., and Tarlow, D.
\newblock Neural networks for modeling source code edits.
\newblock \emph{arXiv preprint arXiv:1904.02818}, 2019.

\end{thebibliography}
\bibliographystyle{icml2020}

\end{document}